%% file: main.tex
\documentclass[acmsmall,nonacm,screen]{acmart}
\usepackage{algorithmic}
\usepackage{graphicx}
\usepackage{textcomp}
\usepackage{xcolor}
\usepackage{color}
\usepackage{listings}
\usepackage[]{hyperref}
\usepackage{mathtools}
\usepackage[ruled, vlined, linesnumbered]{algorithm2e}
\usepackage{circledsteps}
\usepackage{comment}
\usepackage{booktabs}
\usepackage{multirow}
\usepackage{graphicx}
\usepackage{makecell}
\usepackage{titlesec}
\usepackage{subcaption}

\usepackage{pifont}
\newcommand{\X}{$\times$\xspace}
\newcommand{\dffn}{d_{\text{FFN}}}

\newcommand{\showcomments}{yes}
\newcommand\fixme[1]{
    \ifthenelse{\equal{\showcomments}{yes}}{\textcolor{red}{[#1]}}{\ignorespaces}
}
\newcommand\hz[1]{
    \ifthenelse{\equal{\showcomments}{yes}}{\textcolor{red}{[hz: #1~]}}{\ignorespaces}
}
\newcommand\yx[1]{
    \ifthenelse{\equal{\showcomments}{yes}}{\textcolor{blue}{[yx: #1~]}}{\ignorespaces}
}
\newcommand\sj[1]{
    \ifthenelse{\equal{\showcomments}{yes}}{\textcolor{green}{[sj: #1~]}}{\ignorespaces}
}
\newcommand\ns[1]{
    \ifthenelse{\equal{\showcomments}{yes}}{\textcolor{purple}{[ns: #1~]}}{\ignorespaces}
}
\newcommand\yc[1]{
    \ifthenelse{\equal{\showcomments}{yes}}{\textcolor{orange}{[yc: #1~]}}{\ignorespaces}
}
\newcommand\zz[1]{
    \ifthenelse{\equal{\showcomments}{yes}}{\textcolor{violet}{[zhiru: #1]}}{\ignorespaces}
}
\newcommand{\RNum}[1]{\uppercase\expandafter{\romannumeral #1\relax}}
\newcommand{\revise}[1]{\textcolor{black}{#1}}
\newcounter{insight}
\AtBeginDocument{%
  }

\setcopyright{acmlicensed}
\copyrightyear{2018}
\acmYear{2018}
\acmDOI{XXXXXXX.XXXXXXX}

\acmConference[TRETS'24]{TRETS'24}{May 5–8, 2024}{Orlando, FL}
\acmISBN{978-1-4503-XXXX-X/18/06}

\begin{document}

\title{Understanding the Potential of FPGA-Based Spatial Acceleration for Large Language Model Inference}

\author{Hongzheng Chen}
\email{hzchen@cs.cornell.edu}
\affiliation{%
  \institution{Cornell University}
  \country{USA}
}

\author{Jiahao Zhang}
\authornote{Work was done when interning at Cornell.}
\email{jiahao-z19@mails.tsinghua.edu.cn}
\affiliation{%
  \institution{Tsinghua University}
  \country{China}
}

\author{Yixiao Du}
\email{yd383@cornell.edu}
\author{Shaojie Xiang}
\email{sx233@cornell.edu}
\author{Zichao Yue}
\email{zy383@cornell.edu}
\affiliation{%
  \institution{Cornell University}
  \country{USA}
}

\author{Niansong Zhang}
\email{nz264@cornell.edu}
\author{Yaohui Cai}
\email{yc2632@cornell.edu}
\author{Zhiru Zhang}
\email{zhiruz@cornell.edu}
\affiliation{%
  \institution{Cornell University}
  \country{USA}
}

\renewcommand{\shortauthors}{Chen et al.}

\input{sections/0-abstract}

\ccsdesc[500]{Hardware~Hardware-software codesign}
\ccsdesc[500]{Computing methodologies~Neural networks}

\keywords{FPGA, high-level synthesis, large language models, hardware acceleration}


\maketitle

\input{sections/1-introduction}
\input{sections/2-background}
\input{sections/3-modeling}
\input{sections/4-sec-case-study}
\input{sections/5-implementation}
\input{sections/6-experiments}

\input{sections/7-discussions}
\input{sections/8-related-work}
\input{sections/9-conclusion}

\input{sections/ack}

\bibliographystyle{plain}
\bibliography{trets24}


\end{document}

%% file: sections/0-abstract.tex
\begin{abstract}
Recent advancements in large language models (LLMs) boasting billions of parameters have generated a significant demand for efficient deployment in inference workloads. While hardware accelerators for Transformer-based models have been extensively studied, the majority of existing approaches rely on temporal architectures that reuse hardware units for different network layers and operators.
However, these methods often encounter challenges in achieving low latency due to considerable memory access overhead.

This paper investigates the feasibility and potential of model-specific spatial acceleration for LLM inference on FPGAs. Our approach involves the specialization of distinct hardware units for specific operators or layers, facilitating direct communication between them through a dataflow architecture while minimizing off-chip memory accesses. We introduce a comprehensive analytical model for estimating the performance of a spatial LLM accelerator, taking into account the on-chip compute and memory resources available on an FPGA. This model can be extended to multi-FPGA settings for distributed inference. Through our analysis, we can identify the most effective parallelization and buffering schemes for the accelerator and, crucially, determine the scenarios in which FPGA-based spatial acceleration can outperform its GPU-based counterpart.

To enable more productive implementations of an LLM model on FPGAs, we further provide a library of high-level synthesis (HLS) kernels that are composable and reusable.
This library will be made available as open-source.
To validate the effectiveness of both our analytical model and HLS library, we have implemented BERT and GPT2 on an AMD Xilinx Alveo U280 FPGA device.
Experimental results demonstrate our approach can achieve up to 13.4$\times$ speedup when compared to previous FPGA-based accelerators for the BERT model.
For GPT generative inference, we attain a 2.2$\times$ speedup compared to DFX, an FPGA overlay, in the prefill stage, while achieving a 1.9$\times$ speedup and a 5.7$\times$ improvement in energy efficiency compared to the NVIDIA A100 GPU in the decode stage.
\end{abstract}

%% file: sections/1-introduction.tex
\section{Introduction}
\label{sec:intro}
The rapid advancement of Transformer-based large language models (LLMs)~\cite{vaswani2017transformer,rishi2021foundation} has sparked a revolution across a wide range of natural language processing tasks, such as conversational AI~\cite{openai2023gpt4,zheng2023chatbotarena,cohen2022lamda} and code generation~\cite{chen2021codex,li2022alphacode,nijkamp2023codegen}. 
Recent research has brought to light the phenomenon of ``emergence'' in LLMs, where advanced capabilities become evident as the models scale up to billions of parameters~\cite{wei2022emergence,wei2022cot}. 
However, supporting this unprecedented scale poses significant challenges, particularly in terms of computational and memory resources.
At the same time, the increasing use of LLMs in interactive applications like voice assistants and autonomous systems requires hardware accelerators capable of providing both low latency and high energy efficiency~\cite{openai2023gpt4,driess2023palme,pope2022googleinf}.

Recent efforts have primarily focused on improving the performance of LLM inference on GPUs~\cite{aminabadi2022dsinf,fastertransformer2022}, although GPUs are known for their high power consumption and are less suitable for latency-sensitive workloads~\cite{kim2023survey,pope2022googleinf}. There is also an active body of research dedicated to developing specialized hardware accelerators tailored for Transformer models, with several of these efforts using FPGAs as the target platforms~\cite{hong2022dfx,liu2021fqbert,li2020ftrans,peng2021cbbp,hur2023flexrun,qi2021iccad}.

\input{fig-tex/temporal-spatial-arch}

FPGA-based LLM accelerators can be broadly categorized into two architectural paradigms: \emph{temporal architecture} and \emph{spatial architecture}.
In a temporal architecture, a processing engine (PE) capable of performing various tasks is constructed and reused across different layers and models, as shown in Figure~\ref{fig:temporal-spatial-arch}(a). For flexibility, these accelerators typically employ an overlay approach~\cite{hong2022dfx,li2020ftrans,khan2021npe}, where a virtual hardware architecture that executes instructions is ``laid'' on top of the physical FPGA fabric. Overlays provide a more restricted configuration space, allowing for quicker compilation with bitstream reuse across multiple models. However, the use of such temporal architecture requires more frequent off-chip memory access, as intermediate results must be written back to memory. This incurs a cost in terms of both latency and energy consumption that is significantly higher than direct on-chip memory access. Additionally, one could argue that an FPGA overlay will inherently be less efficient than its hardened ASIC counterpart.

In contrast, an FPGA-based spatial architecture typically involves the specialization of distinct PEs for specific operators or layers, facilitating direct communication between them using streaming buffers (e.g., FIFOs or multi-buffers)~\cite{xiang2022heteroflow,wang2021autosa,yaman2017finn,petrica2020dataflowcnn}, as depicted in Figure~\ref{fig:temporal-spatial-arch}(b-c).
This dataflow-style execution substantially reduces off-chip memory accesses and enables the concurrent processing of multiple PEs in a pipelined manner. 
Moreover, the fine-grained programmability of FPGAs allows efficient support of model-specific spatial architectures, which can further leverage efficiency optimizations such as low-bitwidth quantization, custom numerical types, and sparsity~\cite{zhang2021fracbnn,sun2022autovit,yang2023aim,peng2022sparsefpga}. These capabilities can potentially enable highly efficient LLM inference implementations that surpass GPUs, especially in small-batch low-latency scenarios.

However, implementing a spatial architecture for LLM inference presents significant challenges.

\noindent\textbf{Challenge 1: Navigating diverse parallelism in LLMs.}
The generative inference process of LLMs typically consists of two distinct stages: (1) simultaneously processing user prompts and (2) sequentially generating new tokens in an autoregressive manner.
These two stages exhibit significantly different computational and memory characteristics (detailed in \S\ref{sec:modeling}), making it necessary to tailor hardware accelerators for their specific needs.
This challenge cannot be directly addressed by leveraging techniques from the traditional convolutional neural network (CNN) designs~\cite{kim2023survey,zhang2015cnnfpga}.
The large number of parameters and intermediate tensors further complicates the choice between on-chip and off-chip storage.
Additionally, harnessing multiple accelerators for distributed LLM inference adds complexity, particularly when dealing with intricate parallelization schemes~\cite{shoeybi2019megatron,narayanan2019pipedream,hong2022dfx}.

\noindent\textbf{Challenge 2: Lack of standard LLM building blocks in hardware accelerators.}
The rapid evolution of LLM architectures~\cite{rishi2021foundation,openai2023gpt4,touvron2023llama} contrasts with the comparatively slow pace of hardware development.
While a plethora of building blocks for Transformers have been proposed in the software domain~\cite{xFormers2022,dao2022flashattention,kernl2022}, the absence of reusable blocks for hardware accelerator design hampers development progress.
Many frameworks have been designed to automatically map deep learning models to  FPGAs~\cite{zhang2018dnnbuilder,yaman2017finn,fahim2021hls4ml,suhail2023flexcnn,zhang2020dnnexplorer}, but they are constrained to small CNN designs and lack support for complicated Transformer models.
It is also hard to scale their designs to accommodate large models and multi-die FPGAs.

To tackle these challenges,
this paper is to provide a comprehensive set of hardware design considerations for LLMs and try to answer the following question: {\emph{What role can FPGA-based spatial accelerators play in enabling efficient LLM inference?}
We start by conducting an in-depth analysis of the computational and memory requirements associated with each operator within Transformer models across two distinct stages of LLM generative inference -- prefill and decode.
Subsequently, we extend our analysis to reveal the potential benefits of distributed inference using multiple FPGAs.
\textbf{We believe that providing such an analysis, rather than presenting only positive results in selectively chosen settings for an FPGA LLM accelerator, offers more valuable insights to the community.}
To validate the feasibility of our analytical framework, we implement a specific design point and demonstrate its viability.
Leveraging this analytical framework, we employ specific optimizations in HLS to craft each kernel and compose them into a hardware accelerator that achieves the expected performance.
While our primary focus is not to propose a new LLM accelerator architecture, we demonstrate that by using the analytical model, we can create a high-performance design that surpasses previous efforts.
Our major contributions are as follows: 

\begin{itemize}
    \item 
    We introduce an analytical framework that presents the first in-depth analysis of both the advantages and limitations of FPGA-based LLM spatial acceleration.
    This framework not only allows us to estimate the performance of a specific accelerator configuration on a given FPGA device but also provides guidance for designing accelerators for LLM inference.
    
    \item 
    We create a suite of modular and reusable HLS kernels designed for building FPGA-based spatial accelerators for different Transformer models. We plan to open-source this kernel library\footnote{\url{https://github.com/cornell-zhang/allo/tree/main/examples}} and expect it to serve as a valuable resource for benchmarking HLS and FPGA acceleration more broadly.
    
    \item Leveraging our kernel library, we design and implement a range of high-performance FPGA-based LLM accelerators that achieve speedups comparable to previous GPU and FPGA-based accelerators.
    Specifically, for the BERT model, we achieve a 13.4$\times$ speedup over prior FPGA-based accelerators.
    For GPT generative inference, we achieve speedups of 2.2$\times$ and 1.1$\times$ in prefill and decode stages respectively, when compared to DFX, an FPGA-based overlay architecture. 
    Additionally, our accelerator is 1.9$\times$ faster and 5.7$\times$ more energy-efficient than the A100 GPU in the decode stage.
\end{itemize}


%% file: fig-tex/temporal-spatial-arch.tex
\begin{figure}[t]
    
  
    \begin{subfigure}[b]{\linewidth}
        \centering
        \includegraphics[width=0.5\linewidth]{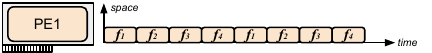}    
        \caption{Temporal architecture (i.e., overlay).}
        \label{subfig:temporal}
    \end{subfigure}
    
    \begin{subfigure}[b]{\linewidth}
        \centering
        \includegraphics[width=0.5\linewidth]{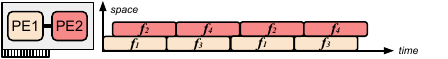}
        \caption{Partially unfolded spatial architecture with two PEs.}
        \label{subfig:partial}
    \end{subfigure}
    
    \begin{subfigure}[b]{\linewidth}
        \centering
        \includegraphics[width=0.5\linewidth]{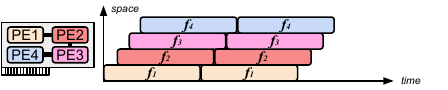}    
        \caption{Fully unfolded spatial architecture with four PEs.}
        \label{subfig:spatial}
    \end{subfigure}
    
    \caption{Temporal and spatial architectures --- \textnormal{PE stands for processing engine; $f_1$-$f_4$ represent different operators in the model.}}
    \label{fig:temporal-spatial-arch}
\end{figure}

%% file: sections/2-background.tex
\section{Background}
\label{sec:background}
This section provides backgrounds on Transformer models and introduces parallelization schemes for LLM inference.


\subsection{Transformer Models}
\label{sec:transformer}
\begin{figure}[!htbp]
    \centering
    \includegraphics[width=0.8\linewidth]{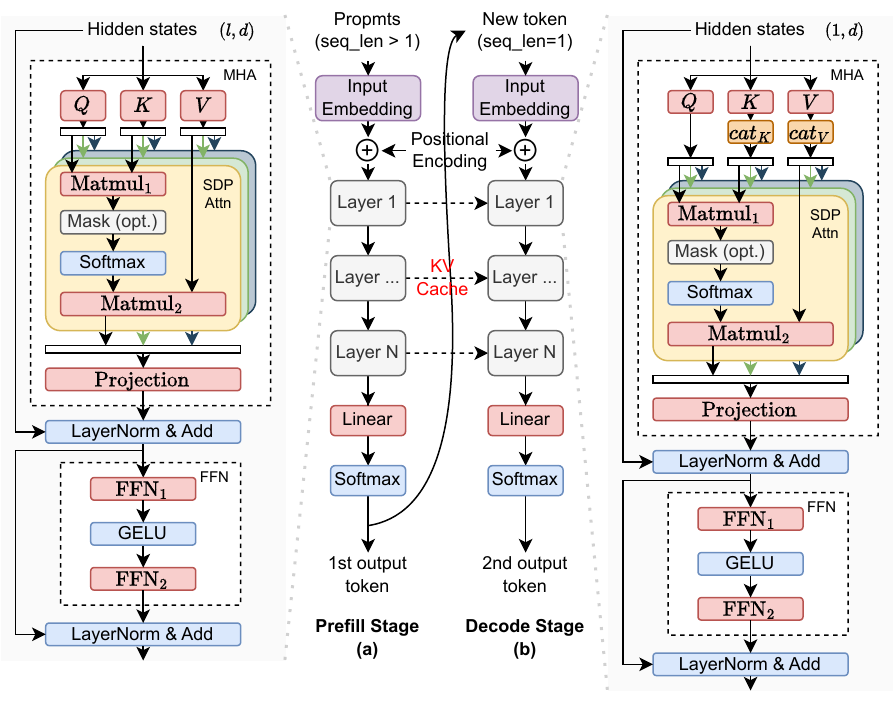}
    \caption{Transformer model. Red blocks represent linear operators, and blue blocks signify non-linear operators.}
    \label{fig:transformer}
\end{figure}

The Transformer model consists of both encoder and decoder blocks~\cite{vaswani2017transformer}.
Recent employment on LLMs mostly uses decoder-only models, which leverage an auto-regressive approach for text generation~\cite{radford2019gpt2,openai2023gpt4,touvron2023llama}.
We will mainly discuss decoder-only models in this paper, but since encoders and decoders share the core building blocks with subtle architectural variances, our approach can also be extended for encoder-only models~\cite{devlin2018bert,lan2020albert,liu2019roberta}.

As illustrated in Figure~\ref{fig:transformer}, generative inference of LLMs has two stages: prefill stage and decode stage~\cite{pope2022googleinf}.
In the prefill stage, the model takes in user prompts, normally a long sequence with $l_{\text{input}}$ tokens, goes through the whole Transformer model, and generates the first token.
In the decode stage, the model takes in the previously generated token and generates $l_{\text{gen}}$ new tokens one at a time in an auto-regressive way.
Since each token depends on the previously generated tokens, the decode stage is purely sequential.

We then go through the detailed model architecture.
The input tokens are first passed into an embedding layer that maps the discrete tokens into high-dimensional continuous representations while incorporating positional encoding for each token.
Subsequently, it generates a tensor (i.e., hidden states) of shape $(l,d)$, where $l$ represents sequence length, and $d$ is the size of hidden dimensions.
We omit the batch dimension to simplify the analysis, focusing solely on single-batch inference in this paper, but our approach can be easily extended to different batch sizes for LLM serving by adding an additional batch dimension~\cite{kwon2023vllm,li2023alpaserve}.

The hidden states then pass through a series of $N$ Transformer blocks.
Each Transformer block consists of two sublayers: a multi-head attention (MHA) module and a feed-forward network (FFN).
Residual connections and layer normalization (LayerNorm) functions are applied between these sublayers, although the specific order and application may vary across different models~\cite{xiong2020preln}.
The MHA module plays a crucial role in capturing token relationships within the input sequence.
The input is initially partitioned into $h$ segments, where $h$ corresponds to the number of attention heads.
To compute the attention scores for each head, the input sequence of length $l$ undergoes three linear projections: query, key, and value.
These projections, which are trainable, yield matrices $Q$, $K$, and $V$ respectively.
Attention scores are then computed using a scaled dot-product (SDP) operator between $Q$, $K$, and $V$, as specified by the formula:
\begin{equation}
\text{Attention}(Q,K,V)=\mathrm{softmax}\left(QK^T\Big/\sqrt{d_k}\right)V\,,
\end{equation}
where $d_k$ is the size of the hidden dimension.
The output from this operator involves $h$ outputs, which are subsequently concatenated and processed through an additional linear projection.
In the prefill stage, the generated $K$ and $V$ tensors will be stored as \emph{KV cache} and later be concatenated before SDP during the decode stage~\cite{pope2022googleinf}.

The FFN module comprises a linear layer followed by a non-linear activation function and another linear layer.
This module transforms the outputs of MHA into embedding matrices, which are then further processed by subsequent Transformer layers.

Finally, the output tensor will go through a softmax function to obtain a distribution.
The model will sample a token from this distribution and feed it into the decode stage.
For encoder-only models, there is only a prefill stage involved, and the distribution will be directly used for different downstream tasks like text classification~\cite{devlin2018bert,liu2019roberta,lan2020albert}.

In this paper, we only focus on analyzing the core Transformer blocks and accelerating them on FPGAs.
Embedding layers and output sampling~\cite{generation_strategies,chen2023speculative} require extensive random memory accesses, which may not be suitable for FPGA acceleration.
Also, they only take a small fraction of overall compute that does not affect the overall latency~\cite{kim2023survey}, so we leave them to execute on CPUs or GPUs as usual.


\subsection{Parallelization Schemes}
\label{sub:parallelism_intro}

\begin{figure}[t]
    \centering
    \includegraphics[width=0.9\linewidth]{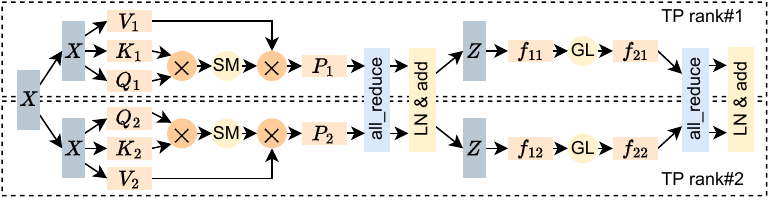}
    \caption{An example of tensor parallelism of a Transformer layer with two devices.
    TP rank is the unique identifier given to a device within a TP group.
    SM is the softmax function, LN is LayerNorm, and GL is the GeLU function.}
    \label{fig:3d-parallelism}
\end{figure}

As model sizes continue to expand, it becomes increasingly common for a single device to be insufficient for accommodating the entire model.
Consequently, the exploration of diverse parallelization schemes within a device and across devices becomes necessary.
Parallelism techniques in deep learning can be roughly classified into data, tensor, and pipeline parallelism, together known as 3D parallelism~\cite{shoeybi2019megatron,narayanan2021megatronv2,korthikanti2022reducing,chen2024slapo}.
During the era of CNNs, data parallelism was the norm, involving the partitioning of input tensors along the batch dimension and their distribution across multiple devices~\cite{li2020ptddp,abadi2016tensorflow}.
DeepSpeed ZeRO~\cite{rajbhandari2020zero} and FSDP~\cite{fsdp} extended data parallelism by proposing a three-stage parallelism strategy that partitions optimizer states, gradients, and parameters to minimize memory usage.
However, this approach may incur high communication overheads during inference.

A more recent parallelism scheme for LLM inference is tensor parallelism (TP)~\cite{li2023alpaserve,aminabadi2022dsinf,shoeybi2019megatron,narayanan2021megatronv2,pope2022googleinf}, which distributes model parameters across multiple devices and conducts explicit collective operations to ensure model correctness.
Megatron-LM~\cite{shoeybi2019megatron} is the first to explore tensor parallelism for Transformer-based models, proving to be efficient in both training and inference due to relatively low communication costs.
As shown in Figure~\ref{fig:3d-parallelism}, tensor parallelism requires two \texttt{all\_reduce} operations inside a Transformer layer to ensure the results are correct.
Our accelerator design also explores tensor parallelism, as detailed in \S\ref{subsub:parallelism_schemes}.

Lastly, pipeline parallelism~\cite{narayanan2019pipedream,yang2021pipemare,narayanan2021pipe2bw} divides the model across network layers.
Multiple layers are grouped into a pipeline stage, and different stages are assigned to different devices.
Pipeline parallelism is typically employed across multiple nodes.
Since both tensor parallelism and pipeline parallelism handle only portions of the network model, they are collectively referred to as model parallelism.
We revisit these parallelization schemes in \S\ref{subsub:parallelism_schemes}.

%% file: sections/3-modeling.tex
\section{Analytical Modeling Framework}
\label{sec:modeling}
In this section, we propose a comprehensive analytical modeling framework aimed at understanding the computational requirements of a Transformer layer.
Our investigation begins by analyzing the compute demands and resource constraints on a single device.
We base our estimations on these constraints.
Finally, we extend the framework to the analysis of multiple devices.


\input{sections/3.1-modeling-macs}

\input{sections/3.2-modeling-constraints}
\input{sections/3.3-modeling-estimation}
\input{sections/3.4-modeling-multi-fpga}

%% file: sections/3.1-modeling-macs.tex
\subsection{Computational Demands}
\label{sub:mac_analysis}
Our first imperative is to calculate the computational demands of the model.
Given that the predominant computation in the Transformer model is general matrix-matrix multiplication (GEMM or Matmul) or general matrix-vector multiplication (GEMV)~\cite{pope2022googleinf,kim2023survey}, we employ the number of multiply–accumulates (MACs) as the proxy metric for quantifying compute requirements of the linear layers, as depicted in Table~\ref{tab:flops}.
\revise{For non-linear layers such as softmax and GeLU functions, they are elementwise operators that can be easily fused with the GEMM kernels in a pipeline design without affecting the final performance. More experimental results are provided in \S\ref{sub:ablation}.}

We denote $X_{(\cdot)}$ as the output tensors from preceding layers, and $W_{(\cdot)}$ represents the weights of corresponding linear layers.
For example, $X_{\text{sm}}$ is the output of the softmax operator.
Our analysis here is restricted to a single batch; hence the tensors only have two dimensions.
We can observe that the computational demand during the prefill stage far surpasses that of the decode stage.
In the prefill stage, the required MACs of the two \texttt{Matmul}s within the SDP are quadratic to the sequence length (i.e., $l^2d$).
Consequently, when input sequences exhibit substantial length, attention layers may extend computation time significantly.
On the contrary, in the decode stage, each operator processes a single token at a time, making the MACs independent of the sequence length except for SDP.

\begin{table}[t]
    \centering
    \caption{MACs of the prefill and decode stages of the linear layers in the Transformer model in Figure~\ref{fig:transformer} --- \textnormal{$l$ denotes input sequence length, $d$ denotes input feature dimension size, and $\dffn$ denotes FFN hidden dimension size.}}
    \label{tab:flops}
    \small
    \begin{tabular}{ccccc}\Xhline{1pt}
        \textbf{Linear Layer}&  \textbf{Abbreviations} & \textbf{Input Matrices} & \textbf{Prefill} & \textbf{Decode}\\\Xhline{1pt}
         Q/K/V linear&  $q$, $k$, $v$ & $XW_Q, XW_K, XW_V$ & $3ld^2$ & $3d^2$\\
         Matmul$_1$&  $a_1$ & $QK^\mathrm{T}$ & $l^2d$ & $(l+1)d$\\
         Matmul$_2$&  $a_2$ & $X_{\text{sm}}V$ & $l^2d$ & $(l+1)d$\\
         Projection&  $p$ & $X_{\text{sdp}}W_{\text{Proj}}$& $ld^2$ & $d^2$\\
         FFN$_1$&  $f_1$ & $X_{\text{mha}}W_{\text{FFN}_1}$ & $ldd_{\text{FFN}}$ & $dd_{\text{FFN}}$\\
         FFN$_2$&  $f_2$ & $X_{\text{act}}W_{\text{FFN}_2}$ & $ldd_{\text{FFN}}$ & $dd_{\text{FFN}}$\\\Xhline{1pt}
    \end{tabular}
\end{table}

%% file: sections/3.2-modeling-constraints.tex
\subsection{Resource Constraints}
We then model the compute and memory resource constraints on an FPGA.
In this section, we assume that one FPGA device can effectively compute at least a single Transformer layer, but our framework can be easily extended to more resource-constrained cases using a similar analysis proposed in \S\ref{sub:multifpga}.

\subsubsection{Compute Resource Constraints.}
\label{subsubsec:compute_resource}
The core computational element for linear operators is the MAC unit.
Let $M_i$ denote the compute power, in terms of the number of MACs per cycle allocated to each matrix multiplication kernel, where $i$ ranges over $q$, $k$, $v$, $a_1$, $a_2$, $p$, $f_1$, and $f_2$, based on the notation in Table~\ref{tab:flops}.
We quantize the matrix multiplication to integer inputs for maximum efficiency, which has been proven to be effective by many recent studies ~\cite{xiao2023smoothquant,dettmers2022llmint8,sheng2020qbert,kim2021ibert}.
Quantization enables single-cycle accumulation.
As a result, one multiply-accumulator (MAC) unit can provide a 1 MAC/cycle throughput with a properly pipelined multiplier.
Therefore, the latency for the $Q$ projection can be calculated as $ld^2/M_{q}$ cycles, considering that the total number of MACs computed in this operator is $ld^2$.

Suppose we want to deploy $C$ Transformer model layers on an FPGA. 
The total MAC units must not exceed the capacity of the device.
Since we employ a dataflow design that unfolds all the layers on-board, the required MAC units are simply the sum of the MAC units for each layer.
This requirement can be expressed as:
\begin{equation}
    \label{eq:mac_constraint}
     \sum{M_i}C < M_{\text{tot}}, {i\in\{q,k,v, a_1, a_2, p, f_1, f_2\}}\,,
\end{equation}
where $M_{\text{tot}}$ represents the total available compute power of an FPGA in terms of MACs per cycle,
which can be obtained from the official data sheets.
\revise{For FPGAs with specialized compute blocks (e.g., AI Engine~\cite{vck5000} and AI Tensor Blocks~\cite{stratix10}), we can convert their compute power to match the frequency of the programming logic, thus obtaining an overall value $M_{\text{tot}}$ for the entire FPGA.
For example, the VCK5000 FPGA~\cite{vck5000} has 400 AI Engines, each of which can compute 128 MACs/cycle at 1GHz.
Therefore, the equivalent compute power at 250MHz is 128$\times$400$\times$1GHz/250MHz, which is 204800 MACs/cycle.}

\subsubsection{Memory Capacity Constraints.}
\label{subsubsec:memory_resource}
The demand for memory capacity stems from a variety of on-chip buffers, including weight buffers for parameters, buffers for $K$ and $V$ matrices, and FIFOs interconnecting different stages.

\textbf{Parameter buffers.}
To optimize an FPGA-based dataflow design, we assume that all the quantized parameters can be accommodated in on-chip or off-chip memory.
Suppose all the linear weights are quantized to $b_W$ bits, and the size of the linear operator $i$ is $s_i$.
The total size of the buffers is $S_{\text{param}}=\sum_{i\in\{q,k,v,p,f_1,f_2\}}s_ib_W=(4d^2+2d\dffn)b_W$ if storing on-chip.
If the parameters are too large to fit in on-chip memory, we can store the parameters in DRAM and tile the parameters with size $M_i$ on-chip, then the total tiled buffer size is $S_{\text{tile}}=\sum_{i\in\{q,k,v,p,f_1,f_2\}}M_ib_W$.
To hide the memory access latency, we need to double buffer those parameters, so the final buffer size of the $i$-th linear operator is $2S_{\text{tile}}$.

\textbf{KV Cache.}
When conducting matrix multiplication, at least one of the matrices' elements must be accessed repeatedly so that a buffer is required.
Given that parameters are already buffered, only the SDP requires buffering for at least one of the input matrices.
In our case, we choose to buffer $K$ and $V$, which will be later passed to the decode stage as the KV cache.
We also double buffer $K$ and $V$ matrices to improve throughput.
The final buffer size is $S_{\text{KV}}=4l_{\max}db_{A}$, where $b_{A}$ is the bitwidth of the activation and $l_{\max}$ is the maximum sequence length supported by the model.
Notice KV cache can also be tiled on-chip, which can leverage a similar analysis above.


\textbf{FIFOs.}
The intermediate results between linear operators flow in FIFOs since the linear operators sequentially access them.
For the initial residual connection, we assume that the input tensors are fetched from off-chip memory to obviate the need for additional buffering. 
However, for the second residual connection related to the FFN, it is necessary to use an intermediate buffer to store the projection's activation $X_{\text{act}}$ before the FFN. 
This buffer simultaneously serves as a bypass path.
To avoid deadlock, the buffer must possess sufficient capacity to store $X_{\text{act}}$.
We simply create a FIFO of size $ldb_{A}$ to store it.
For other FIFO connections, we assume a FIFO depth of $s$ and one FIFO connecting each layer in Figure~\ref{fig:transformer}, so the total FIFO size is equal to $S_{\text{FIFO}}=16sb_{A}+ldb_A$.

In summary, the memory capacity constraint is expressed as:
\begin{equation}
\label{eq:sram_constraint}
\begin{aligned}
    S_{\text{param}}C &< DRAM_{\text{tot}}\,,\\
    \sum S_iC &< SRAM_{\text{tot}}\,, i\in\{\text{tile},\text{KV},\text{FIFO}\}\,,
\end{aligned}
\end{equation}
if the parameters are stored off-chip.
$DRAM_{\text{tot}}$ and $SRAM_{\text{tot}}$ are the total available off-chip and on-chip memory.

\subsubsection{Memory Port Constraints.}
\label{subsubsec:memory_port}
Besides memory capacity, we also need to consider constraints on memory ports in a highly paralleled design.
For matrix multiplication, if different MAC units
work in parallel, they will visit the weight/result buffers simultaneously, hence contending for memory ports.
This issue can be addressed by either partitioning the buffer, effectively offering more memory ports; or packing data to create wider elements, subsequently reducing the number of memory ports required.

\textbf{SRAM resources.}
The on-chip SRAM resources of FPGAs are typically organized as blocks. 
Each block has a fixed capacity and may support configurable bitwidth.
For example, on AMD UltraScale+ FPGAs, there are two types of SRAM resources: \emph{Block RAM (BRAM)} and \emph{Ultra RAM (URAM)}.
BRAM blocks can be configured to 1$\times$36 Kb block or 2$\times$18 Kb blocks, with two read and write ports each.
URAM blocks are 288 Kb with one read and one write port.
The port width of the BRAM block is flexible; it can be configured to 1, 2, 4, 9, 18, 36, or 72 (in 36 Kb mode) bits, while the port width of the URAM block is fixed at 72 bits.
Similar to BRAM and URAM, Intel FPGAs have M20K and eSRAM with different configurable port widths.

\textbf{Memory blocks needed without data packing.}
To begin with, we analyze the port constraints without data packing.
In this case, to eliminate the port contention, different MAC units may need different memory ports.
Consider the linear operator $i$ with the size of $s_i$ with $M_i$ MAC units working in parallel, each loaded weight may feed multiple MAC units due to intrinsic data reuse in GEMM. 
We use $r_i$ to represent the data reuse factor (number of MAC units sharing the loaded weight).
Therefore, the weight buffer needs to be partitioned into $M_i/r_i$ parts.
If we store all the weight buffers on-chip, then the number of $b_{W}$-bit elements in each partition is $s_i/(M_i/r_i)$.
However, $b_W$ may not fully occupy one memory word as the memory bitwidth can only take limited options. 
We introduce the effective bit width, $b_{BRAM}$, to be the smallest memory bitwidth larger than $b_W$.
Let $S_{BRAM}$ be the total capacity (in bits) of one memory block,
we can deduce the total number of memory blocks for one linear operator:
\begin{equation}
    R_i=\left\lceil \frac{s_ib_{BRAM}}{M_i/r_i \times S_{BRAM}} \right\rceil \times M_i/r_i\,.
\end{equation}
If the parameters are loaded from off-chip memory and we only store a tile of the weight on-chip, then $s_i$ is simply $M_i$, and $R_i$ also becomes $M_i$ as $b_{BRAM}\ll S_{BRAM}$.
Since we need to double buffer those parameters, the final buffer size of the $i$-th linear operator is $2M_i$.
Notice the $k$ and $v$ layers need to be double-buffered, so the required BRAM also doubles in these two layers.
We can obtain the total required BRAM as below:
\begin{equation}
\label{eq:port_constraint}
\sum_{i\in\{q,k,v,p,f_1,f_2\}}CR_i + 2C(R_{a_1} + R_{a_2}) < Mem_{\text{tot}}\,.
\end{equation}


\textbf{Memory blocks needed with data packing.}
Data packing can alleviate the strain on memory port contention by consolidating multiple narrow data into a single, wider data element.
This process allows multiple MAC units to access data from the same memory port.
We consider packing data into $b_{pack}$ bits for the linear weights, and we have $b_{pack}=kb_W$.
Again, we denote $b_{BRAM}$ as the smallest memory bitwidth larger than $b_{pack}$.
We need to partition $M_i/r_i$ MAC units to $M_i/r_i/k$ parts, and each partition has $\lceil s_i/k\times b_{BRAM}/(M_i/r_i/k)\rceil$ bits.
Therefore, the total number of memory blocks needed is:
\begin{equation}
\label{eq:port_packed_constraint}
R_i=\left\lceil \frac{s_ib_{BRAM}}{M_i/r_i\times S_{BRAM}} \right\rceil \times \frac{M_i/r_i}{k}\,.
\end{equation}


\subsubsection{Memory Bandwidth Constraints.}
\label{subsub:memory_bandwidth}
If the parameters are stored off-chip, we need to consider the impact of off-chip memory bandwidth.
Similar to \S\ref{subsub:memory_packing}, we use $r_i$ to denote the data reuse factor of a linear operator with $M_i$ MAC units.
Effectively, $M_i/r_i$ weights must be loaded from off-chip memory per cycle to feed the MAC units, requiring a bandwidth of:
\begin{equation}
\label{eq:bandwidth}
    B_i = b_W\times M_i/r_i \times freq\,,
\end{equation}
where $freq$ is the achieved frequency of FPGA.
If the total required bandwidth, $\sum_i C{B_i} (i \in \{q,k,v,p,f_1,f_2\})$, exceeds the maximum device bandwidth, the inference becomes bandwidth bound.
Notice this bandwidth requirement needs to be analyzed for each operator individually if the data loading requires accessing multiple DDR or HBM channels.

%% file: sections/3.3-modeling-estimation.tex
\subsection{Performance Estimation}
In this section, we estimate the overall latency based on the constraints and conduct work balancing for the dataflow.
\begin{figure}[!htbp]
    \centering
    \includegraphics[width=0.6\linewidth]{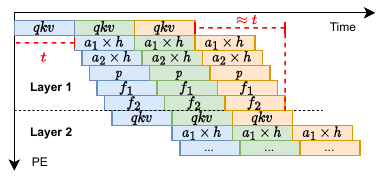}
    \caption{Pipeline diagram. Different colors stand for different input samples.
    Different blocks stand for different linear operators which also constitute the pipeline stages.
    $h$ is the number of attention heads.}
    \label{fig:pipeline}
\end{figure}

\subsubsection{Latency Estimation.}
\label{subsubsec:latency}
We construct the pipeline diagram as shown in Figure~\ref{fig:pipeline}.
As mentioned in \S\ref{subsubsec:memory_resource}, since we need to store the $K$ and $V$ values after the linear operators, there is an implicit synchronization point between the $q$/$k$/$v$ operator and the latter SDP and FFN parts.
The computation of them cannot be overlapped.
Notice the $q$/$k$/$v$ operator can be performed in parallel since they do not have any dependencies.
\revise{After $k$ and $v$ have been fully calculated, the subsequent computations of SDP and FFN can be greatly overlapped.
This is because these operations do not need to wait for all the results to perform the next operation.
The results of the previous operation can be directly streamed into the next operation as input.}
Moreover, since different Transformer layers share the same architecture, their computation can also be overlapped without waiting for the result of the previous layer.

Suppose the Transformer model has $N$ layers in total.
Since we have $C$ layers on one FPGA, it needs to iterate $N/C$ times to process the whole model.
We can calculate the latency of different stages, and the overall latency is the maximum latency of these stages (which defines the initiation interval of the pipeline) times the number of iterations, i.e.,
\begin{align}
    \label{eq:latency_prefill}
    T_{\text{prefill}} &= \frac{1}{freq}\frac{N}{C}\left(\frac{ld^2}{M_k}+C\max\left(\frac{ld^2}{M_k},\frac{l^2d}{M_{a_1}},\frac{ld\dffn}{M_{f_1}},T_{\text{mem}}\right)\right)\,,\\
    \label{eq:latency_decode}
    T_{\text{decode}} &= \frac{1}{freq}\frac{N}{C}\left(\frac{d^2}{M_k}+C\max\left(\frac{d^2}{M_k},\frac{(l_{\max}+1)d}{M_{a_1}},\frac{d\dffn}{M_{f_1}},T_{\text{mem}}\right)\right)\,,
\end{align}
where the first term inside the parentheses is the latency of the $q$/$k$/$v$ linear operator (i.e., $t$ in Figure~\ref{fig:pipeline}).
$T_{\text{mem}}$ is the off-chip memory access latency, which can be calculated based on Equation~\eqref{eq:bandwidth}.

\subsubsection{Work Balancing.}
As the overall latency is determined by the slowest stage in the dataflow, we can balance the execution time of each stage; hence we have
\begin{align}
    &\frac{ld^2}{M_{q,k,v,p}}=\frac{l^2d/h}{M_{a_{1},a_{2}}}h=\frac{ld\dffn}{M_{f_{1},f_{2}}}\\
    \label{eq:m}
    \implies& M=M_{q,k,v,p}=d/l M_{a_{1},a_{2}}=d/\dffn M_{f_{1},f_{2}}\,,
\end{align}
where $M$ is defined as the global compute power in MACs/cycle.
Finally, Equation~\eqref{eq:latency_prefill} can be simplified to 
\begin{equation}
T_{\text{prefill}}=\frac{1}{freq}N\left(1+\frac{1}{C}\right)\frac{ld^2}{M}\,,
\end{equation}
which shows the overall latency with work balancing.
We can obtain the latency for the decode stage using a similar analysis.

To derive the optimal $M$ for a given model, we devise a linear search algorithm to identify the maximum available $M$ based on the constraints in Equations~\eqref{eq:mac_constraint}, \eqref{eq:sram_constraint}, and \eqref{eq:port_packed_constraint}.
Notice the optimal $M$ represents an upper bound of the compute power.
In practice, we also need to consider the routing issue to adjust the actual achievable $M$ as discussed in \S\ref{sec:xcel}.


%% file: sections/3.4-modeling-multi-fpga.tex
\subsection{Distributed Inference}
\label{sub:multifpga}
As a single FPGA may not be sufficient to process some extremely large models, we next extend our modeling to multiple FPGAs.
We first characterize the communication cost between two FPGAs and discuss the impact of different parallelism schemes.

\subsubsection{Communication.}
Various methods exist for facilitating inter-FPGA communication, including communication through the host, PCI-E Peer-to-Peer (P2P), and on-device Ethernet.
We mainly consider the third approach since it does not necessitate orchestration from the host and provides higher bandwidth compared to other alternatives.
For example, the AMD Alveo U280 FPGA provides two QSFP ports~\cite{qsfp}, each capable of carrying 100 Gb/s Ethernet data over optical fibers, which ensures robust and high-speed inter-FPGA communication.
Most of the time, we cannot fully utilize the network bandwidth and need to pay for the package header overheads.
Suppose the theoretical network bandwidth between two FPGA devices is $B$ bits per second (bps), and the efficiency of the network is $\alpha$, so we can have the effective bandwidth as $\alpha B$, where $\alpha$ can be obtained through network benchmarking.


\subsubsection{Parallelization Schemes.}
\label{subsub:parallelism_schemes}
As mentioned in \S\ref{sub:parallelism_intro}, we have various parallelization schemes when considering multiple devices.
We first analyze tensor parallelism (TP).
\revise{As shown in Figure~\ref{fig:3d-parallelism}, the parameters of the linear operations are partitioned across different devices. For example, suppose the weight parameters of the two FFN layers $f_1$ and $f_2$ are $A$ and $B$, then we can partition $A$ along its column and partition $B$ along its row, and obtain
\[\sigma(ZA)B=\sigma\left(Z\begin{bmatrix}A_1 & A_2\end{bmatrix}\right)\begin{bmatrix}B_1\\B_2\end{bmatrix}=\sigma(ZA_1)B_1+\sigma(ZA_2)B_2\,,\]
where $\sigma$ is the GeLU function.
Therefore, apart from partitioning $A$ and $B$, we need to insert an all-reduce operation to aggregate the partial results on each device to ensure correctness.
The partitioned parameters will be stored on different devices.
For example, $A_1$ will be on the first FPGA, and $A_2$ will be on the second FPGA.
A similar partition scheme can be applied for MHA, and we refer the readers to~\cite{shoeybi2019megatron} for more details.
}

\begin{figure}[!htbp]
    \centering
    \includegraphics[width=0.6\linewidth]{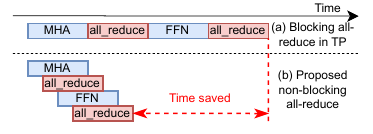}
    \caption{Blocking and non-blocking all-reduce in TP. The latency of different stages is not drawn to scale.}
    \label{fig:tp}
\end{figure}

Based on this partition scheme, TP requires two all-reduce operations within one Transformer layer.
However, these communicative operations are implemented in a blocking way.
Figure~\ref{fig:tp}(a) shows the subsequent FFN module needs to wait for the completion of the all-reduce process before it can conduct computation~\cite{wang2023overlap}.
Notice that the all-reduce operation only involves fetching results from other devices and adding the result to its local tensor.
Given that the output of MHA is a sequential stream, we can perform elementwise addition in a non-blocking manner.
As soon as the kernel receives enough data, it can initiate data transfer to other devices without waiting for the remaining data to be computed.
This leads to substantial synchronization time savings as shown in Figure~\ref{fig:tp}(b).

Since the size of the output tensor of MHA and FFN are both $ld$, the communication time for one all-reduce is
\begin{equation}
    T_{\text{comm}}=ldb_A/(\alpha B)\,.
\end{equation}

\revise{As we have already implemented dataflow inside a device, pipeline parallelism (PP) essentially extends the dataflow to $p_2$ devices with a tensor of size $ld$ communicated in between.
Here, we only split the pipeline between two Transformer layers so the results of the previous device can be directly streamed to the next device in the same PP group.
Notice TP and PP can be combined to conduct model inference~\cite{narayanan2021megatronv2}}, and the latency of Equation~\eqref{eq:latency_prefill} becomes
\begin{equation}
    \label{eq:multilatency}
    \small
    T_{\text{prefill}} = \frac{1}{freq}\frac{N}{p_2C}\left(\frac{ld^2}{p_1M_k}+p_2C\max\left(\frac{ld^2}{p_1M_k},\frac{l^2d}{p_1M_{a_1}},\frac{ld\dffn}{p_1M_{f_1}}, T_{\text{mem}},T_{\text{comm}}\right)\right)\,,
\end{equation}
where $p_1$ and $p_2$ are the size of a TP group and a PP group~\cite{shoeybi2019megatron}.
Additionally, the memory requirements of Equations~\eqref{eq:sram_constraint} and \eqref{eq:port_constraint} need to be divided by $p_1$ to satisfy the constraints of multiple devices.

Notice we only discuss two basic parallelism schemes for Transformer models.
Some recent works may partition the sequence dimension and leverage reduce-scatter and all-gather to reduce the overheads of all-reduce~\cite{narayanan2021megatronv2,korthikanti2022reducing}.
The communication time can be similarly analyzed, and we will not discuss them here.
The optimal parallelism scheme on multiple devices~\cite{pope2022googleinf,colin2022unity,zheng2022alpa,miao2022galvatron,xie2022optimalplacement} is out of the scope of this paper, and we will leave it as future works.

%% file: sections/4-sec-case-study.tex
\section{Case Study}
\label{sec:case_study}
In this section, we leverage actual hardware configurations to estimate the model performance using our analytical framework and provide insights for LLM accelerator design.

\subsection{Overview of Workloads and Hardware}
\label{sub:workload_hardware}
Table~\ref{tab:models} lists several widely used models that we choose for performance estimation. 
Such models include BERT-base~\cite{devlin2018bert}, a representative encoder-only model,
GPT2~\cite{radford2019gpt2}, the only open-sourced model in the GPT family,
LLaMA-7B~\cite{touvron2023llama,touvron2023llama2}, an open-sourced model trained by Meta,  
and Vicuna-13B~\cite{chiang2023vicuna}, the best non-commercial model on Chatbot Arena~\cite{zheng2023chatbotarena}.
\begin{table}[t]
    \centering
    \caption{Models used in \S\ref{sec:case_study} and \S\ref{sec:exp}.}
    \label{tab:models}
    \small
    \begin{tabular}{ccccccc}\Xhline{1pt}
        \multirow{2}{*}{\textbf{Model}} & \multirow{2}{*}{\textbf{Type}} & \multirow{2}{*}{\textbf{\# of params}} & \textbf{\# Layers} & \textbf{\# Heads} & \textbf{Hidden} & \textbf{FFN size}\\
        &  & & \textbf{$N$} & \textbf{$h$} & \textbf{Size $d$} & \textbf{$d_{FFN}$}\\\Xhline{1pt}
        BERT~\cite{devlin2018bert} & Encoder & 110M & 12 & 12 & 768 & 3072 \\
        GPT2~\cite{radford2019gpt2} & Decoder & 355M & 24 & 16 & 1024 & 4096 \\
        LLaMA2~\cite{touvron2023llama2} & Decoder & 7B & 32 & 32 & 4096 & 11008\\
        Vicuna~\cite{chiang2023vicuna} & Decoder & 13B & 40 & 40 & 5120 & 13824\\
        \Xhline{1pt}
    \end{tabular}
\end{table}

To see the performance differences between FPGAs and GPUs, we pick and list several representative devices in Table~\ref{tab:device}.
For FPGAs, Alveo U280 and Agilex 7 are two FPGAs that are widely used in cloud servers but not specially optimized for AI workloads;
Versal VCK5000 and Stratix 10 NX FPGAs are designed for accelerating AI applications with specialized hardware units such as AI Engine (AIE)~\cite{xilinxAIE} or Tensor Blocks~\cite{stratix10}.
Versal VHK158 is the latest FPGA with HBM2e released by AMD in 2023.
For GPUs, RTX 2080Ti is a high-end GPU designed for personal usage with a similar process node and release date to U280;
A100 is the most deployed GPU in data centers to conduct LLM training and inference~\cite{openai2023gpt4}.

\begin{table*}
    \centering
    \caption{Summary of FPGA and GPU devices.}
    \label{tab:device}
    \small
    \resizebox{\linewidth}{!}{
    \begin{tabular}{llllllll}\Xhline{1pt}
     & \multicolumn{3}{c}{\textbf{AMD Xilinx FPGA}} & \multicolumn{2}{c}{\textbf{Intel FPGA}} & \multicolumn{2}{c}{\textbf{Nvidia GPU}} \\
     & \makecell[l]{\textbf{Alveo}\\\textbf{U280~\cite{u280}}} & \makecell[l]{\textbf{Versal}\\\textbf{VCK5000~\cite{vck5000}}} & \makecell[l]{\textbf{Versal}\\\textbf{VHK158~\cite{vhk158}}}& \makecell[l]{\textbf{Stratix 10}\\\textbf{NX 2100~\cite{stratix10}}} & \makecell[l]{\textbf{Agilex 7}\\\textbf{AGM039~\cite{agilex7}}} & \makecell[l]{\textbf{GeForce}\\\textbf{RTX 2080 Ti}} & \textbf{Tesla A100} \\\Xhline{1pt}
     \textbf{Process Node} & TSMC 16nm & TSMC 7nm & TSMC 7nm & Intel 14nm & Intel 7nm & TSMC 12nm & TSMC 7nm\\
     \textbf{Release Date} & 2018 & 2022 & 2023 & 2020 & 2022 & 2018 & 2021\\
     \textbf{Thermal Design Power} & 225W & 225W & 180W & 225W & 225W & 250W & 300W\\
     \textbf{Peak Throughput} & 24.5 INT8 TOPS & 145 INT8 TOPS & 56 INT8 TOPS & 143 INT8 TOPS & 88.6 INT8 TOPS & 14.2 TFLOPS & 312 TFLOPS\\
     \textbf{Specialized Blocks} & - & \makecell[l]{400$\times$\\AI Engine} & - & \makecell[l]{3960$\times$\\AI Tensor Block} & - & \makecell[l]{544$\times$\\Tensor Cores} & \makecell[l]{432$\times$\\Tensor Cores}\\
     \textbf{DSP/CUDA Cores} & 9024 & 1968 & 7392 & - & 12300 & 4352 & 6912\\
     \textbf{BRAM18K/M20K} & 4032 & 967 & 5063 & 6847 & 18960 & - & -\\
     \textbf{URAM/eSRAM} & 960 & 463& 1301 & 2 & - & - & -\\
     \makecell[l]{\textbf{On-chip Memory}\\\textbf{Capacity}} & 41MB & 24MB & 63.62MB & 30MB & 46.25MB & 5.5MB & 40MB\\
     \makecell[l]{\textbf{Off-chip Memory}\\\textbf{Capacity}} & \makecell[l]{8GB HBM2 \&\\32GB DDR} & 16GB DDR & \makecell[l]{32GB HBM2e \&\\32GB DDR}& 16GB HBM2 & 32GB HBM2e & 11GB DDR & 80GB HBM2e\\
     \makecell[l]{\textbf{On-chip Memory}\\\textbf{Bandwidth}} & \makecell[l]{460GB/s \&\\38GB/s} & 102.4GB/s & \makecell[l]{819.2GB/s \&\\102.4GB/s} & 512GB/s & 820GB/s & 616GB/s & 1935GB/s\\
     \Xhline{1pt}
    \end{tabular}
    }
\end{table*}

\subsection{Single-Device Performance Estimation}
We first conduct experiments on a single device with pre-trained BERT-base and GPT2 models from HuggingFace Hub~\cite{wolf2019huggingface}.
For GPU baselines, we run the models with PyTorch 2.0~\cite{paszke2019pytorch} and CUDA 11.7, and measure the performance on both RTX 2080Ti and A100 GPUs.
The host machine runs an Intel Xeon Silver 4114 CPU at 2.20GHz with 40 cores.
We follow the common practice to measure the out-of-the-box FP16 performance~\cite{aminabadi2022dsinf,fastertransformer2022,radford2019gpt2}.

In this section, we only estimate the performance of FPGAs using the proposed framework in \S\ref{sec:modeling}, and in \S\ref{sec:exp} we will evaluate the actual performance on FPGAs.
The quantization scheme is 4-bit weight and 8-bit activation (W4A8) configurations for FPGA estimations, unless otherwise noted.

\subsubsection{Latency of Different Stages.}
\begin{figure*}
    \centering
    \includegraphics[width=\linewidth]{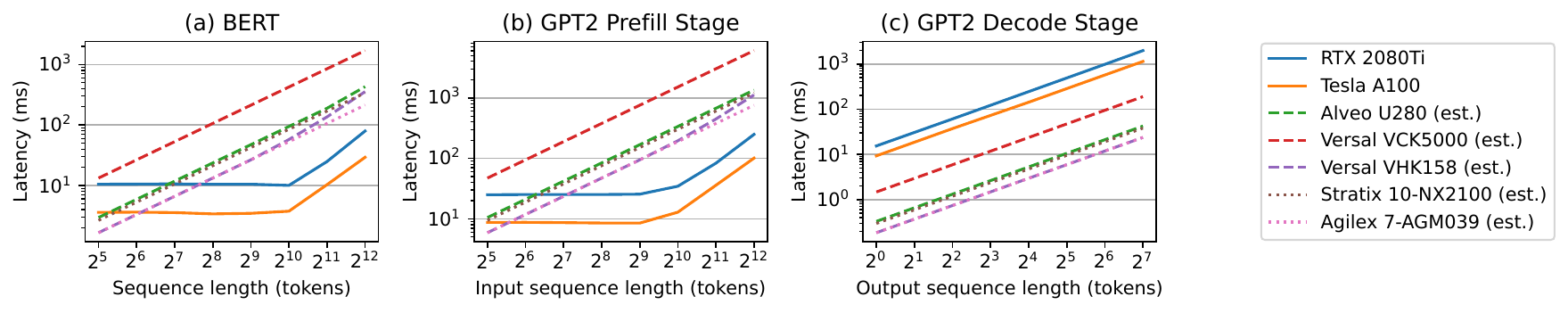}
    \caption{Latency estimation of BERT and GPT2 on different FPGAs.
    GPU results are obtained from actual profiling.}
    \label{fig:bert_gpt_latency}
\end{figure*}
Based on the constraints of Equations~\eqref{eq:mac_constraint}, \eqref{eq:sram_constraint}, and \eqref{eq:port_packed_constraint}, we can calculate the latency one device can achieve at the maximum compute power $M$ (defined in Equation~\eqref{eq:m}).

\revise{From Figure~\ref{fig:bert_gpt_latency}(a) and (b), we observe that the 2080Ti and A100 GPUs maintain an almost constant curve when the sequence length is less than 1024.
This behavior is primarily attributed to the high kernel launch overheads for hundreds of CUDA kernels in PyTorch, which overshadow the computation time.
However, when the sequence length exceeds 1024, GPUs become computation-bound, resulting in higher latency.
In contrast, the latency of FPGAs increases linearly with the sequence length, as described in Equation~\eqref{eq:latency_prefill}, and demonstrates significantly longer latency when the sequence length is large (e.g., 512).}
It is worth noting that even when making use of state-of-the-art AI-optimized FPGAs like the Stratix 10 NX and Versal VCK5000, the situation does not improve significantly.
This is because these modern FPGAs are equipped with DDR or older versions of HBM, which makes parameter loading from off-chip memory the bottleneck.
Even though these FPGAs have highly efficient computation blocks, the compute units have to wait for the memory to fetch data, resulting in suboptimal performance.
\revise{Moreover, many FPGAs struggle to achieve both high memory bandwidth and compute power simultaneously.
Consequently, for VHK158, performance deteriorates when the sequence length reaches 4096, as it shifts from being memory-bandwidth bound to compute-bound.
Further scaling the sequence length larger than 4096 may lead to out-of-memory for both FPGAs and GPUs, and only A100 can handle such large sequence lengths, so the latency is not plotted on the figure.}

The decode stage is on the contrary, as shown in Figure~\ref{fig:bert_gpt_latency}(c).
When the sequence length is small, GPUs suffer from underutilizing the compute resources, and FPGAs can achieve a significantly lower latency compared to GPUs.
The latency of RTX2080 and A100 GPUs increases since the memory access takes control compared to computation in the decode stage.
Although FPGAs are also bounded by off-chip memory bandwidth, they can always perform better than GPUs since the computation is small when the sequence length is equal to one.
Therefore, FPGAs can easily achieve GPU-level performance even with a small $M$.

\stepcounter{insight}
\textbf{Insight \RNum{\theinsight}: Existing FPGAs are inferior in the compute-intensive prefill stage but can outperform GPUs in the memory-intensive decode stage.}

\begin{figure}[t]
    \centering
    \includegraphics[width=0.65\linewidth]{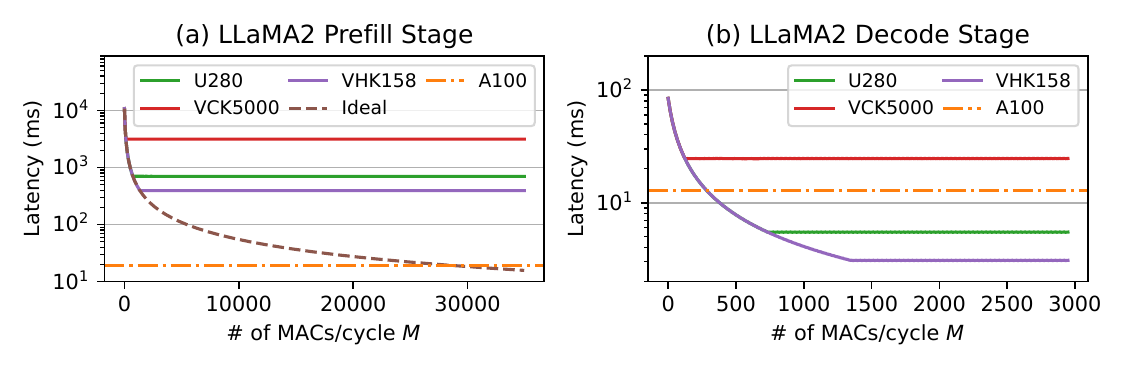}
    \caption{Latency estimation of LLaMA2 model.
    The sequence length is set as 128, and the W4A8 quantization scheme is used in this experiment.
    GPU results are obtained from actual profiling.}
    \label{fig:llama}
\end{figure}

To further investigate what constrains the performance of FPGAs, we conduct an analysis on the LLaMA2 model by varying different $M$ and observing the changes in latency.
As shown in Figure~\ref{fig:llama}, the VCK5000 FPGA exhibits the smallest off-chip memory bandwidth, which leads it to reach a latency plateau rather quickly.
Conversely, the VHK158 FPGA has the largest off-chip memory bandwidth, so it can achieve the lowest latency in both prefill and decode stages.
Moreover, we include the curve of ideal FPGA performance in Figure~\ref{fig:llama} to assess the compute power required to attain A100-level performance.
Based on this estimation, we need around 30,000 MACs/cycle in order to achieve the A100-level performance in the prefill stage, assuming no memory bandwidth constraints.
This is achievable by those AI-optimized FPGAs, which can conduct a large number of MACs efficiently.
On the contrary, for the decode stage, once an FPGA has enough memory bandwidth, such as U280, it can reach the A100-level performance easily.

\stepcounter{insight}
\textbf{Insight \RNum{\theinsight}: The prefill stage requires large compute power $M$ to achieve the GPU-level performance, while the decode stage only requires a small $M$.}

\subsubsection{Quantization Schemes.}
\label{subsub:memory_packing}
\begin{figure}[t]
    \centering
    \includegraphics[width=\linewidth]{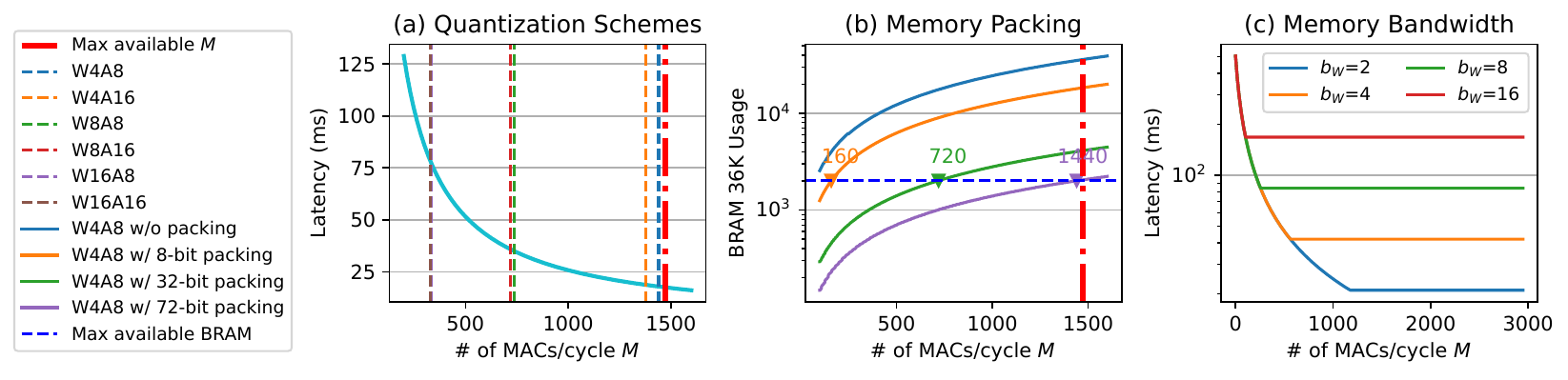}
    \caption{(a) Impact of different quantization techniques on GPT2 prefilling stage on U280.
    The sequence length is set as 128.
    The cyan line shows the theoretical latency under different $M$ without memory bandwidth constraints.
    Thin dashed lines depict the maximum $M$ constrained by available BRAM resources.
    (b) Impact of memory packing in the W4A8 setting.
    (c) Impact of different weight quantization schemes on memory bandwidth and overall latency.}
    \label{fig:bitwidth}
\end{figure}

We then investigate the impact of different quantization schemes and memory packing.
We consider quantizing the weight parameters to $x$ bits and the activation to $y$ bits (abbreviated as W$\{x\}$A$\{y\}$).
As shown in Figure~\ref{fig:bitwidth}(a), the red dashed line depicts the maximum available MACs/cycle on-board, which is calculated based on Equation~\eqref{eq:mac_constraint}.
Different quantization schemes may have different requirements on BRAM usage constrained by Equation~\eqref{eq:sram_constraint}.
W4A8 is the scheme that can almost fully utilize the compute resources.
W8A8 and W16A16 require more memory resources, resulting in lower performance since the computation is bound by the limited BRAM resources on-board.
Also, we can see quantizing the weights gives the most benefits, but quantizing activation only gives little benefit ($M$ does not change a lot under the same weight bitwidth), which is due to the fact that we employ a dataflow architecture and do not require large buffers to store the intermediate tensors on-board.

\stepcounter{insight}
\textbf{Insight \RNum{\theinsight}: Weight quantization is necessary for reducing memory usage, while activation quantization only has limited benefit.}

\subsubsection{Memory Packing.}
Next, we further consider the impact of memory packing under the W4A8 setting.
As shown in Figure~\ref{fig:bitwidth}(b), if we do not conduct memory packing, it even cannot satisfy the memory port constraint (Equation~\eqref{eq:port_constraint}) when $M$ is small (blue curve).
This is because a large number of partitioned arrays require more BRAMs, and many BRAMs are not fully utilized causing a large waste of resources.
The orange curve shows packing two \texttt{int4} elements to \texttt{int8}, and we can achieve a small $M$ under the resource constraint since the number of partitioned arrays is reduced.
The green curve packs 9$\times$\texttt{int4} elements to \texttt{int36}, and it can achieve more than four times of $M$ compared to the \texttt{int8} packing.
The purple curve packs 18$\times$\texttt{int4} elements to \texttt{int72}, and the curve can almost intersect with the red line before intersecting with the blue line, which means it reaches the maximum DSP constraint on-board (Equation~\eqref{eq:mac_constraint}).
This study shows that it is important to pack the parameters to reduce on-chip memory usage.

\stepcounter{insight}
\textbf{Insight \RNum{\theinsight}: Memory packing can efficiently reduce the required BRAMs to store the tensors.}

\subsubsection{Memory Bandwidth.}
Lastly, we investigate how quantization impacts the required memory bandwidth.
As shown in Figure~\ref{fig:bitwidth}(c), the low-bit weight quantization can significantly alleviate the demands of off-chip memory access.
By reducing the volume of data needed in each cycle, it can achieve a larger compute power $M$, thus leading to a better performance.
In particular, quantizing the model to a 2-bit representation yields a performance boost exceeding an order of magnitude when compared to a 16-bit weight quantization scheme.
Recent research~\cite{zhao2023atom,chee2023quip} has demonstrated that 4-bit or even 2-bit quantization can be implemented without compromising model accuracy, which makes efficient LLM deployment on FPGAs possible.

\stepcounter{insight}
\textbf{Insight \RNum{\theinsight}: Low-bit weight quantization can further help alleviate the demands of off-chip memory access.}

\subsection{Multi-Device Performance Estimation}
\begin{figure}[!htbp]
    \centering
    \includegraphics[width=0.7\linewidth]{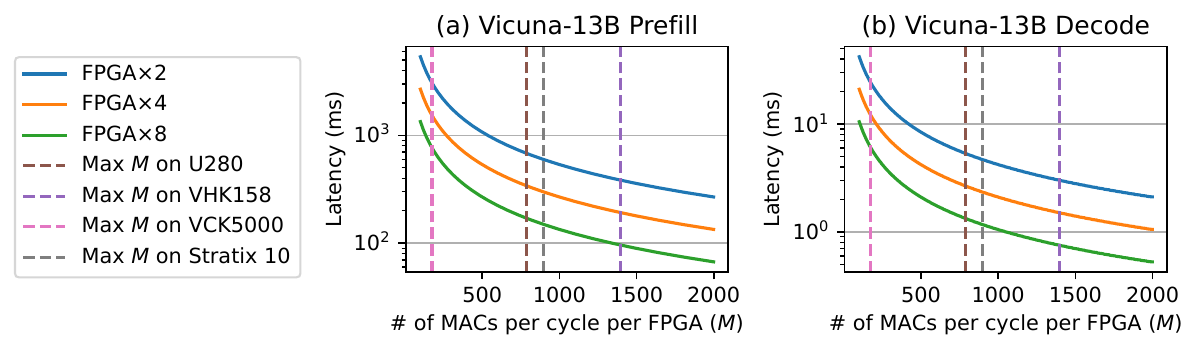}
    \caption{Latency estimations of Vicuna-13B model on multiple FPGAs.}
    \label{fig:multifpga}
\end{figure}
For multiple devices, we use the Vicuna-13B model to estimate the performance of 2, 4, and 8 FPGAs based on our analytical model.
As shown in Figure~\ref{fig:multifpga},
the latency can scale well when the number of devices increases.
Since we employ a non-blocking communication scheme in our dataflow design as discussed in \S~\ref{sub:multifpga}, communication will not be the bottleneck of the design.
Multiple FPGAs can reduce the number of required MACs on each device, but cannot increase the number of available MACs on an FPGA, so the performance is still limited by the maximum available resources on-board and the off-chip memory bandwidth.
For the decode stage, leveraging two FPGAs can already reduce the inference latency of the Vicuna-13B model to less than 10ms based on the estimation.

\stepcounter{insight}
\textbf{Insight \RNum{\theinsight}: Multiple FPGAs help reduce overall latency under the same $M$ on each device.}

%% file: sections/5-implementation.tex
\section{Implementations}
\label{sec:implementation}
In this section, we describe the kernel implementation and accelerator design to show how to efficiently achieve the design points in the analytical framework.

\input{sections/5.1-kernels}

\input{sections/5.2-accelerator}

%% file: sections/5.1-kernels.tex
\subsection{HLS Kernels}
\label{subsec:kernels}
This section is not intended to propose new HLS kernels.
Instead, we explore the efficient ways to reach the maximum available $M$ on FPGAs and implement the standard kernels as a library with parameterized quantization support, which is reusable across different models (e.g., BERT and GPT models implemented in \S\ref{sec:exp}).

\noindent\textbf{Linear Operators.}
Linear operators are the key operators in Transformer models because they are ubiquitous and compute-intensive.
There are two types of linear operators in the Transformer layers, activation-weight matrix multiply (A-W GEMM) with bias, and activation-activation (A-A GEMM) matrix multiply.
A-W GEMM includes the projection layers and the linear layers in the FFN, with weights buffered on-chip and activations streamed from an input FIFO;
A-A GEMM includes the two GEMM operators in the SDP attention, with one of the input activations stored in a double buffer and the other one streamed.

\begin{figure}[!htbp]
    \centering
    \includegraphics[width=0.7\linewidth]{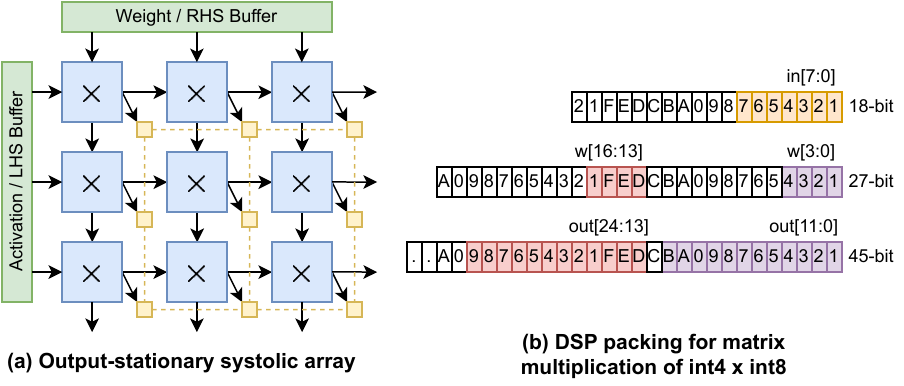}
    \caption{Systolic array and DSP packing. The yellow blocks in the systolic array represent output buffers.}
    \label{fig:sa}
\end{figure}

We adopt an output-stationary systolic array architecture to implement the A-W and A-A GEMM process engines.
As shown in Figure~\ref{fig:sa}(a), the systolic array is a 2-D array of MAC units of shape $m_1\times m_2$ with FIFOs connecting different MAC units.
The number of MAC units of linear operator $i$ is actually $M_i$ defined in \S\ref{sub:mac_analysis}.
We mainly discuss the A-W GEMM in the following, and the idea can also be applied to the A-A GEMM.
The input activation from the previous layer will be first buffered in an activation buffer.
After $m_1$ elements are buffered, they will be streamed into the systolic array together with the $m_2$ weights.
There is also a fully partitioned output buffer on-chip that allows the outputs to directly write back.

Each MAC unit can be implemented with a single DSP block and can provide one-MAC-per-cycle throughput.
Based on the discussion in \S\ref{subsub:memory_packing}, we adopt the W4A8 quantization scheme for our accelerator design, which maximizes the utilization of available resources.
As a result, the matrix multiplications involve either \texttt{int4} by \texttt{int8} or \texttt{int8} by \texttt{int8} operations, which are too large for LUTs, thus relying primarily on DSPs or specialized compute blocks (e.g., AIE~\cite{xilinxAIE}).
In AMD FPGAs, the DSP48E2 hard blocks can support 18-bit by 27-bit multiplication and accumulation~\cite{u280}, enabling the packing of two multiplications into one slice for a W4A8 quantized model to save DSP resources and achieve a larger $M$.
Figure~\ref{fig:sa}(b) shows the bit packing method for 4-bit by 8-bit integer multiplications.
One activation is filled into the lower 8 bits of the 18-bit DSP input, and two weights are filled into 0-to-3 and 13-to-16 bit positions of the 27-bit DSP input to avoid overlapping results.
Finally, the two multiplication results are extracted by bit-slicing the 45-bit DSP result.
Notice that since the DSP output is wide enough, we can also pack two 8-bit by 8-bit integer multiplications into one DSP slice by further offsetting the second weight and output.
With DSP packing, we can easily double $M$ to achieve higher performance but with much fewer DSPs.


\noindent\textbf{Non-Linear Operators.}
Since quantizing non-linear operators can lead to a significant degradation in model accuracy~\cite{sheng2020qbert,xiao2023smoothquant}, and these non-linear operators are not the bottleneck of the design, we directly implement the floating-point version of the operators in HLS.
Specifically, we buffer a row of elements for softmax and LayerNorm functions, which requires conducting reduction along the last dimension.
Consequently, this approach eliminates the need to wait for the complete results for the computation of these non-linear operators and effectively prevents dataflow stalling.

%% file: sections/5.2-accelerator.tex
\begin{figure*}
    \centering
    \includegraphics[width=\linewidth]{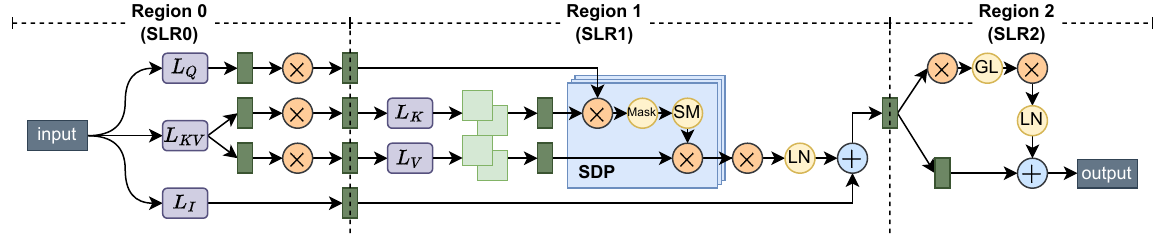}
    \caption{Overall dataflow architecture of a single Transformer layer that uses post-LayerNorm scheme~\cite{radford2019gpt2}.
    SDP denotes scaled dot-product.
    Orange nodes denote the GEMM kernels.
    Yellow nodes are the non-linear kernels, including softmax (SM), LayerNorm (LN), and GELU (GL).
    Green rectangles represent the FIFOs between kernels, and purple rectangles are the data loaders.}
    \label{fig:arch}
\end{figure*}

\subsection{Accelerator Design}
\label{sec:xcel}
We integrate the proposed kernels in \S\ref{subsec:kernels} to create a high-performance hardware accelerator.
The overall architecture of our proposed accelerator is depicted in Figure~\ref{fig:arch}.
Different operators including linear and non-linear operators are connected with FIFOs, except that the KV cache discussed in \S\ref{subsubsec:memory_resource} leverages double buffers.
This design reads input tensors from off-chip memory and stores the results back to memory after each layer.
Intermediate activation values are directly streamed to the next operator.
\revise{Initially, all parameters are stored in DRAM, and data loaders $L_{(\cdot)}$ are responsible for loading data from DRAM or streaming buffers.
Storing parameters off-chip reduces the number of FPGA devices needed and avoids the need to build different bitstreams for different network layers.
After completing a layer, the accelerator fetches new parameters from the host to the device for the subsequent layer.
Since data loading is hidden from the computation, overall latency remains unaffected.}
Given the contemporary trend of multi-die FPGA designs~\cite{guo2021autobridge,du2022hisparse}, explicit dataflow partitioning becomes necessary to meet timing requirements.
Our target device, Alveo U280 FPGA~\cite{u280}, has three chip dies called super logic regions (SLRs).
Consequently, we partition the dataflow into three regions to confine each region fully inside SLR.
\revise{According to our placement constraints, AMD Vitis toolchain will automatically insert AXI Register Slice IPs to pipeline SLR crossing.}


We leverage the proposed analytical framework to guide our accelerator design.
Since typical Transformer models have $\dffn=4d$~\cite{devlin2018bert,radford2019gpt2} and $l<d$, according to work balancing of Equation~\eqref{eq:m}, we have $M_{q,k,v,p} = M$, $M_{a_1, a_2} < M$, and $M_{f1, f2} = 4M$.
A straight-forward division is to put the PEs for $q$, $k$, $v$, SDP, and $p$ on SLR0, $f1$ on SLR1, and $f2$ on SLR2 so that each SLR roughly contains $4M$ MAC units.
However, we observe that scaling up the linear operators in FFN poses significant challenges to timing closure. 
Among various configurations of systolic arrays we tested, the maximum capacity of one SLR at 250 MHz is three of 8~\X16 systolic arrays; a single 16~\X16 one fails timing. 
Therefore, we only leverage 8~\X8 and 8~\X16 systolic arrays for simplicity. 
\revise{We also explore using LUT-based multipliers as they provide greater flexibility for placement compared to DSPs.
However, the presence of additional inter-LUT wires results in a much lower frequency (191 MHz) compared to the DSP-based multipliers.}
To minimize the number of SLR crossings, we put $q$, $k$, and $v$ on SLR0 and use 8~\X16 systolic arrays, which also ensures a relatively low latency for the first stage based on Equation~\eqref{eq:latency_prefill}.
MHA and the $p$ projection are on SLR1, with $a_1$ and $a_2$ using 8~\X8, and $p$ using 8~\X16 systolic arrays.
$f_1$ and $f_2$ operators on SLR2 using 8~\X16 systolic arrays.
Therefore, it can still form a relatively balanced 3:2:2 resource utilization ratio for linear operators.

%% file: sections/6-experiments.tex
\section{Evaluation on FPGAs}
\label{sec:exp}


In this section, we implement two design points studied in \S\ref{sec:case_study} to validate the feasibility of our framework.
We first describe our experimental setup and perform evaluation on a single FPGA.

\subsection{Experiment Setup}
We test the publicly available BERT and GPT2 models listed in Table~\ref{tab:models}.
We conduct post-training quantization and use the W4A8 quantization scheme for BERT and the W8A8 scheme for GPT, which are prevalent settings in nowadays LLM inference~\cite{zhao2023atom,xiao2023smoothquant}.

\begin{figure*}
    \includegraphics[width=0.5\linewidth]{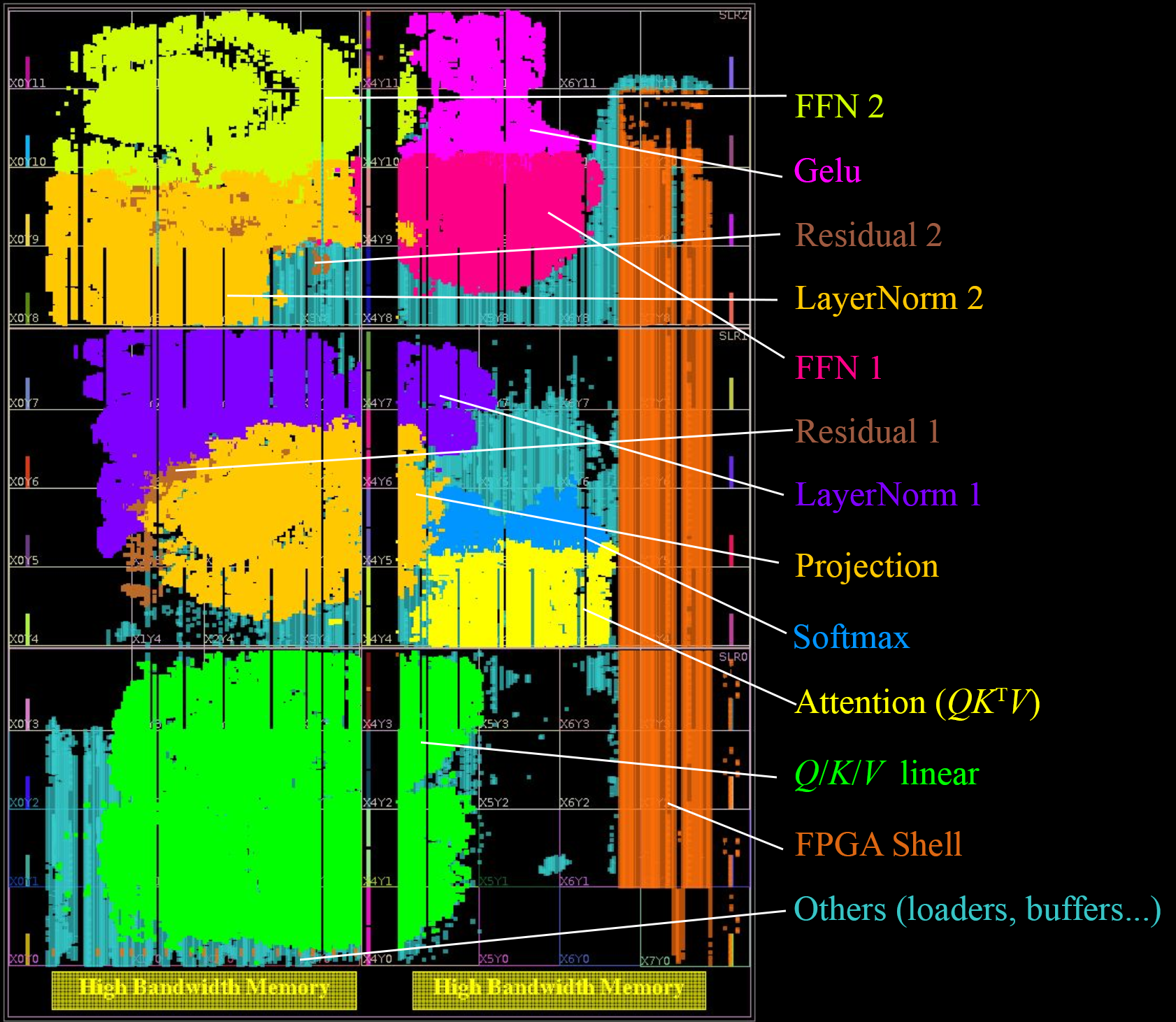}
    \caption{Physical layout of the implemented spatial accelerator.}
    \label{fig:device-layout}
\end{figure*}

We implement and run the actual design on an Alveo U280 FPGA~\cite{u280} with $M=256$ for the kernels.
This FPGA has 4032 BRAM18K blocks, 9024 DSP slices, 2.6M flip-flops, 1.3M LUTs, and 960 URAM blocks.
It has three SLRs with an almost equal amount of resources.
All the kernels are implemented in C++ with Vitis HLS v2022.1~\cite{vitis_hls_2022}, and synthesized with Vivado backend toolchains.
\revise{Figure \ref{fig:device-layout} shows the final device layout of the implemented accelerator.}
We use OpenCL with Xilinx RunTime (XRT) for hardware execution and Xilinx Board Utility (xbutil) for computing power measurements.
The environment for GPU experiments is listed in \S\ref{sub:workload_hardware}, and NVIDIA system management interface (nvidia-smi) is used for measuring GPU power.
Notice the quantized models on GPUs are slower than the FP16 models, as the quantization methods normally leverage fake quantization and lack high-performance GPU kernels to support efficient inference.
Therefore, we directly compare our accelerator with the best FP16 GPU results.
The FPGA on-board results match the outputs from the quantized model in PyTorch and are able to achieve the same accuracy.
The latency results are the average across fifty runs.

\subsection{On-Board Evaluation}

\begin{table*}
\centering
\caption{Experimental results compared with other FPGA-based accelerators. Sequence lengths are set as 512.}
\label{tab:bert_results}
\resizebox{\linewidth}{!}{
{
\begin{tabular}{l l l l l l l l l l l l}\Xhline{1pt}
\textbf{Name}  & \textbf{Device} & \makecell[l]{\textbf{Freq}\\\textbf{(MHz)}}  & \textbf{Quantization} &
\makecell[l]{\textbf{Latency (ms)}\\\textcolor{gray}{[\textbf{Est. (\S\ref{sec:modeling})}]}}
& \makecell[l]{\textbf{Throughput}\\\textbf{(samples/sec)}} & \textbf{Speedup} & \textbf{BRAM} & \textbf{DSP} & \textbf{FF} & \textbf{LUT} & \textbf{URAM}\\
\Xhline{1pt}
\textbf{Ours} & U280 & 245 & W4A8 & 26.01 \textcolor{gray}{[24.07]} & \underline{\textbf{38.45}} & - & 389 (19\%) & 1780 (20\%) & 653K (25\%) & 569K (44\%)& 111 (12\%)\\
 - SLR0 & -& -& -& 4.86 & -& -& 130 (19\%) & 482 (17\%) & 200K (23\%) & 167K (38\%) & 3 (1\%)\\
 - SLR1 & -& -& -& 14.63 & -& -& 136 (20\%) & 590 (19\%) & 240K (28\%) & 212K (49\%) & 50 (16\%)\\
 - SLR2 & -& -& -& 19.81 & -& -& 123 (18\%) & 708 (23\%) & 213K (25\%) & 191K (44\%) & 58 (18\%)\\\hline
FQ-BERT~\cite{liu2021fqbert} & ZCU111 & 214 & W4A8 & 95.16 & 10.51 & \textbf{3.66$\times$} & 679 (31\%)& 3287 (77\%)& 201K (24\%)& 190K (45\%)& N/A\\\hline
TRAC~\cite{patrick2022trac} & ZCU106 & 200 & Fixed8 & 347.48 & 2.88 & \textbf{13.36$\times$} & 181 (29\%) & 1379 (80\%)& 128K (28\%)& 126K (55\%)& N/A\\
\Xhline{1pt}
\end{tabular}
}}
\end{table*}
\begin{figure*}
    \centering
    \includegraphics[width=\linewidth]{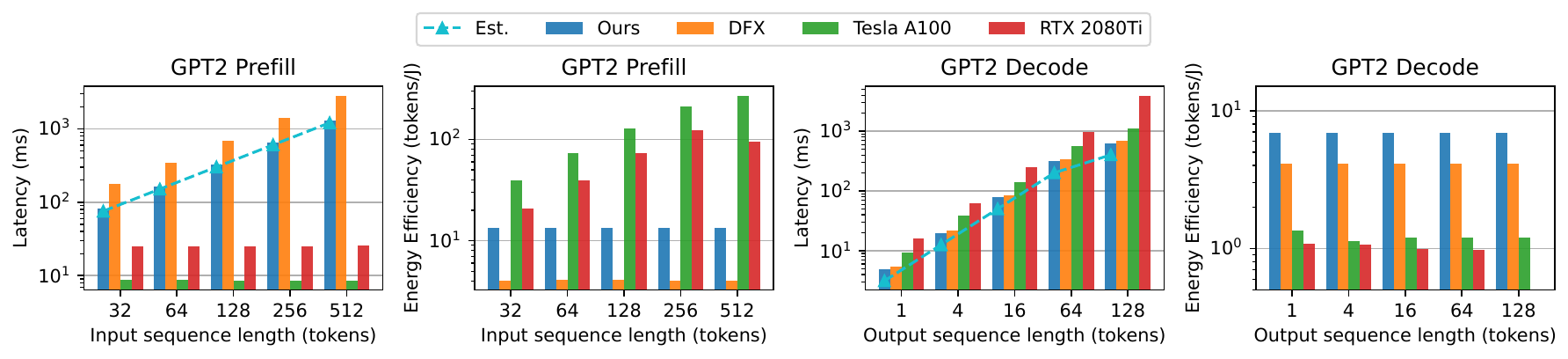}
    \caption{Latency and energy efficiency of GPT2 model on different devices.
    The GPU results are obtained following the same setting in \S\ref{sec:case_study}.}
    \label{fig:gpt2_results}
\end{figure*}

We first compare our BERT accelerator with FQ-BERT~\cite{liu2021fqbert} and TRAC~\cite{patrick2022trac}, two FPGA-based accelerators for the BERT model.
To make a fair comparison, we employ the same W4A8 quantization precision with FQ-BERT and assess the model accuracy using the CoLA dataset~\cite{devlin2018bert}, a widely-used language understanding task.
The fp16 model achieves an accuracy of 56.84\%, while our W4A8 quantized model attains 56.76\% accuracy.
We only compare the performance of the core encoder layers, and \revise{report the best on-board latency results of the baselines obtained from the original papers.
Both baselines use a sequence length of 128, so we scale their latency results to match our sequence length of 512.
FTrans~\cite{li2020ftrans} also targets BERT-variant models, but it does not provide frequency and sequence length in the paper, so we cannot make a direct comparison.}
From Table~\ref{tab:bert_results},
we can see our spatial architecture delivers a much lower latency and a higher throughput with a much lower DSP usage.
Even though our evaluation device is not exactly the same as the baselines, our throughput improvement still surpasses the resource increment of FF and LUT.
Specifically, our accelerator is 3.66$\times$ faster than FQ-BERT~\cite{liu2021fqbert} and 13.36$\times$ faster than TRAC~\cite{patrick2022trac}.
\revise{Compared to temporal architectures, our spatial architecture can efficiently overlap the execution of the operators in the model and eliminate most of the data movement overhead.
In our design, one layer can start operation once the previous layer finishes computing one tile of the feature map, which typically takes only tens of cycles.
In contrast, temporal architectures need to wait for the entire tensor to be produced, which may take hundreds of cycles.
Therefore, even if the per-layer latency of spatial architectures is longer, the end-to-end latency can still be significantly lower than the temporal architectures employed by FQ-BERT and TRAC.}
Furthermore, our analytical framework precisely predicts the performance of the accelerator with less than 2ms of differences, showing the practicality of our approach.

We next design an accelerator for the GPT2 model.
We support importing quantized models from different quantization frameworks~\cite{xiao2023smoothquant,zhao2023atom,chee2023quip}.
Specifically, we export the W8A8 model from SmoothQuant~\cite{xiao2023smoothquant} and achieve 62.2\% on the LAMBADA dataset~\cite{denis2016lambada}, whereas the FP16 model demonstrates an accuracy of 65.0\%.
We compare our GPT accelerator with the state-of-the-art GPT accelerator, DFX~\cite{hong2022dfx}, which employs a temporal architecture with an instruction set and uses the same U280 FPGA device for on-board evaluation.
On average, we are 2.16$\times$ and 1.10$\times$ faster than DFX in the prefill and decode stage respectively.
This is because our spatial architecture overlaps the computation and greatly eliminates off-chip memory access.
We can also see our estimations in \S\ref{sec:modeling} align closely with the actual performance, achieving a 92\% accuracy for the prefill stage.
For the decode stage, the estimated latencies are lower than the actual results, which is mainly because the initial interval between two operators is not significantly smaller than the execution time of one stage, contributing to a notable increase in latency.

We also include the GPU results in \S\ref{sec:case_study} for a more comprehensive evaluation.
As shown in Figure~\ref{fig:gpt2_results}, neither DFX nor our design performs well during the prefill stage compared to GPUs that have more compute resources to exploit the abundant parallelism.
Notably, the latency of FPGAs in the prefill stage increases linearly, while the GPU ones almost remain constant as the model does not fully utilize GPUs.
For the decode stage, the situation is reversed.
FPGA-based accelerators are more efficient than GPUs, and our accelerator can achieve a 1.85$\times$ speedup and is 5.69$\times$ more energy efficient compared to the A100 GPU.
This is because the generation of each token is fully sequential, and GPUs cannot leverage their abundant parallelism, and suffer from extensive memory access overheads.
On the contrary, our dataflow accelerator eliminates most of the off-chip memory accesses and overlaps the compute as much as possible.
Thus, we can achieve a better performance compared to GPUs, aligning with our estimation results in \S\ref{sec:case_study}.
Notice the U280 FPGA only uses a 16nm process while the A100 GPU has a more advanced 7nm process node based on the data in Table~\ref{tab:device}, but we can still achieve higher speedup, demonstrating the efficiency of our spatial accelerators.
It also indicates the potential of further optimizing our HLS design and scaling it up to achieve even higher performance.

\subsection{Ablation Study}
\label{sub:ablation}
\revise{We begin by examining the latency of different SLRs.
As shown in rows 3-5 in Table~\ref{tab:bert_results}, the overall latency of the BERT accelerator is nearly the sum of the latency of SLR0 and SLR2, aligning with the pipeline diagram in Figure~\ref{fig:pipeline}.
Moreover, the computation in SLR1 largely overlaps with that of SLR2 due to the fully pipelined design.
The resource usage across different SLRs is also similar, resulting in a balanced design.}

We further investigate the efficiency of our kernel functions.
We conduct experiments on our template systolic array function with the AutoSA-generated systolic array, which is a highly optimized state-of-the-art systolic array implementation~\cite{wang2021autosa}.
From Table~\ref{tab:sa}, we can see our implementation achieves the same level of performance compared to AutoSA.
While maintaining the same DSP usage, the resource usage of our kernel function is much smaller than AutoSA.
\revise{Since the prediction of a single GEMM kernel is accurate and can achieve the theoretical performance, our analytical model can precisely predict the performance of spatial accelerators when combining multiple linear operators.}
The presence of latency-predictable kernels as foundational components plays a pivotal role in this predictability.
Additionally, our function offers enhanced customizability, accommodating varying sizes and the choice of different quantization schemes.

Moreover, employing DSP packing further reduces the DSP usage, allowing one DSP to handle two MAC operations within a single cycle, a feature not supported in AutoSA.
This experiment shows the efficiency of our kernels, facilitating the development of high-performance Transformer accelerators.
\begin{table}[!htbp]
\centering
\caption{Latency and resource usage of our systolic array library function. Results are directly derived from the HLS report in 300MHz.
The GEMM kernel is extracted from the first FFN layer in the BERT-base model with size $(512,768)\times(768,3072)$.
We use a 16~\X16 systolic array to calculate the \texttt{int8} GEMM.
The theoretical peak performance without DSP packing is $(512\times 768\times 3072)/(16\times 16)$ cycles $\times 3.33$ ns/cycle = $15.71$ ms.
}
\label{tab:sa}
\resizebox{\linewidth}{!}{
\begin{tabular}{ccccccc}\Xhline{1pt}
 & \textbf{Latency (ms)} & \textbf{Effective $M$} & \textbf{BRAM} & \textbf{DSP} & \textbf{FF} & \textbf{LUT}\\\Xhline{1pt}
Ours (w/o DSP packing) & 15.73 & 256 & 0 (0\%) & 256 (2\%) & 88284 (3\%) & 168190 (12\%)\\
Ours (w/ DSP packing) & 15.73 & 128 & 0 (0\%) & \textbf{128 (1\%)} &  79969 (3\%) & 244439 (18\%)\\
AutoSA~\cite{wang2021autosa} & 15.71 & N/A & 514 (12\%) & 256 (2\%) & 100138 (3\%) & 244032 (18\%)\\\Xhline{1pt}
\end{tabular}
}
\end{table}

\revise{Lastly, we analyze the performance of the non-linear operators.
As shown in Table~\ref{tab:nonlinear}, we observe that the softmax operator in the MHA module incurs the highest latency, primarily due to the need to compute the exponential function.
Since these operators are elementwise and only require a row of data to start the computation, they can be easily fused with the preceding linear operators in the pipeline, thereby not significantly impacting the overall latency.
For instance, the combined latency of a GEMM kernel (10.77ms) and the softmax operator (6.67ms) greatly exceeds the latency of SLR1 in Table~\ref{tab:bert_results} (14.63ms$/$300MHz$\times$245MHz), indicating substantial overlap between the softmax operator and other operators.
Again, these ablation studies show that considering only the linear operators in the analytical framework is sufficient to achieve an accurate latency estimation.
}
\begin{table}[!htbp]
\centering
\caption{Performance and resource usage of non-linear operators in our kernel library.
Kernel sizes are set to match those of the BERT model in Table~\ref{tab:bert_results}.
Results are directly derived from the HLS report in 300MHz.}
\label{tab:nonlinear}
\small
{
\begin{tabular}{cccccc}\Xhline{1pt}
\textbf{Operator} & \textbf{Latency (ms)} & \textbf{BRAM} & \textbf{DSP} & \textbf{FF} & \textbf{LUT}\\\Xhline{1pt}
Softmax & 6.67 & 8 & 38 & 4835 & 7447\\
LayerNorm & 0.85 & 20 & 80 & 18751 & 12746\\
GeLU & 0.67 & 0 & 256 & 26193 & 16472\\\Xhline{1pt}
\end{tabular}
}
\end{table}

%% file: sections/7-discussions.tex
\section{Discussion}
\label{sec:discussion}
\revise{In the previous sections, we provide details of the analytical framework and prove that it can achieve high accuracy compared to the latency of actual implementation. However, our framework may have limitations when analyzing the overlay designs or compressed models with sparsity, which requires changes in the resource and latency estimation.}
In this section, we will delve into several unanswered questions and open challenges.

\noindent\textbf{AI-Optimized FPGAs.} 
In \S\ref{sec:case_study}, we demonstrate the potential of leveraging FPGAs with specialized compute engines to accelerate LLMs.
Although AIEs and tensor blocks provide massive compute power~\cite{xilinxAIE,stratix10}, the memory layout and bandwidth requirements remain undiscovered. 
Future FPGAs for AI workloads should provide enough memory bandwidth and efficient on-chip interconnect to facilitate the local data movements in a spatial architecture.
Moreover, these specialized hardware blocks usually adopt a unique programming model with custom compilation flows.
It is still an open question whether existing practices for programming those hardware blocks enable efficient execution of Transformer models.


\noindent\textbf{Timing Closure on Multi-Die FPGAs.} 
We encounter timing problems in partitioning and scaling our design in \S\ref{sec:xcel}. 
In general, it is hard to adequately explore the design space of multi-die partitioning and scaling. 
There are automated frameworks~\cite{guo2021autobridge,moazin2023pasta} to generate floorplanning constraints, but they are currently not expressive enough to capture the various data movement schemes (e.g., residual connection, multi-head splitting) within Transformer models.
We hope similar tools for Transformers could be derived from our analytical framework to speed up the design closure. 


\noindent\textbf{Heterogeneous Deployment.} 
Nowadays, data centers are increasingly heterogeneous, with CPUs, GPUs, and FPGAs available at scale~\cite{caulfield2016microsoft,chung2018brainwave,narayanan2021megatronv2}.
Therefore, it is possible to leverage the advantages of different hardware to accelerate Transformer models.
For example, GPUs are good for the GPT prefill stage due to their high compute power; FPGAs can achieve low-latency decode stage with customized spatial architecture.
The key challenge is to build a distributed system that efficiently manages hundreds of heterogeneous devices.
We hope our analysis on resource constraints, latency, and scaling could assist future deployment and evaluation of LLMs in a heterogeneous and distributed environment.


%% file: sections/8-related-work.tex
\section{Related Work}
\label{sec:related_work}
\noindent\textbf{FPGA-Based Transformer Accelerators.}
Most of the prior works on hardware accelerators leverage temporal or overlay architecture with one FPGA~\cite{liu2021fqbert, li2020ftrans, qi2021iccad, hur2023flexrun, khan2021npe,peng2021cbbp,li2021nmta,qi2021bmc}. 
Their performance usually suffers from frequent data movements of intermediate results. 
DFX~\cite{hong2022dfx} explores using multiple FPGAs to accelerate GPT2 inference, but it is still an overlay design.
Some research has delved into software-hardware co-design to optimize the attention kernel~\cite{zhang2021attn_codesign}.
These endeavors often lack in-depth analysis on resource utilization and cannot be easily generalized to other kernels.

\noindent\textbf{Quantization on LLMs.} 
Initial investigations~\cite{xiao2023smoothquant, dettmers2022llmint8, yao2022zeroquant} demonstrate lossless 8-bit quantization for LLMs.
Subsequent studies~\cite{frantar2022gptq, lin2023awq, yao2022zeroquant, yuan2023rptq, kim2023squeezellm, zhao2023atom} keep lowering the bit width;
the latest advancements reveal that 2-bit~\cite{chee2023quip} and even 1-bit (binary) quantization~\cite{zhang2023binarized} are adequate for an accurate LLM.
While these approaches offer valuable insights, our focus remains orthogonal to quantization, as we illustrate optimization techniques and provide high-performance building blocks for deploying quantized LLMs on FPGAs.

\noindent\textbf{HLS Kernel Libraries.}
Despite the existence of kernel libraries for accelerating Transformer models on GPUs~\cite{dao2022flashattention,wolf2019huggingface,xFormers2022}, the hardware domain has seen only a handful of initiatives in this regard.
AMD provides Vitis HLS library~\cite{vitis_hls_library,vitis_ai} that only has basic kernel-level examples without comprehensive designs tailored for Transformer models.
\revise{TRAC~\cite{patrick2022trac} attempts to provide an HLS-based Transformer library, but its kernel performance is unpredictable, and it exclusively focuses on the BERT model using a temporal architecture.}
Some frameworks map deep learning frameworks to FPGAs~\cite{zhang2018dnnbuilder,fahim2021hls4ml,yaman2017finn,blott2018finnr,zhang2020dnnexplorer}, but can only handle small CNN designs and do not cater to LLMs.
More recent tools allow hardware design using Python~\cite{lai2019heterocl,xiang2022heteroflow,huang2021pylog,yehpca2022scalehls,debjit2022dac}, but are still general-purpose and require hardware engineers to construct and optimize kernels from scratch.
Our work provides a Transformer kernel library designed for dataflow implementations and demonstrates their composability in constructing high-performance hardware accelerators.

%% file: sections/9-conclusion.tex
\section{Conclusion}
\label{sec:conclusion}
In this paper, we propose an analytical framework for large language models and point out the bottlenecks and potential optimizations across the prefill and decode stages in the generative inference.
To verify the feasibility of our framework, we provide a reusable HLS kernel library to quickly compose Transformer kernels into different LLMs that can achieve the expected performance.
Based on these proposed kernels, we design FPGA-based spatial accelerators for both BERT and GPT models and achieve high performance and energy efficiency on par with high-end GPUs.
By offering insights into performance bottlenecks, a suite of reusable kernels, and a high-performance accelerator, we propel the deployment of LLMs for real-world applications while pushing the boundaries of hardware innovation.

%% file: sections/ack.tex
\begin{acks}
This work was supported in part by ACE, one of the seven centers in JUMP 2.0, a Semiconductor Research Corporation (SRC) program sponsored by DARPA and NSF Awards \#2007832, \#2019306, and \#2118709.
We would like to thank anonymous reviewers, Keisuke Kamahori, and Zihao Ye for providing insightful feedback.
We also thank Jiajie Li, Jie Liu, and Zhanqiu Hu for their contributions to the initial LLM modeling and benchmarking.
\end{acks}